\documentclass[10pt,twocolumn,letterpaper]{article}

\usepackage{iccv}              % To produce the CAMERA-READY version
\definecolor{iccvblue}{rgb}{0.21,0.49,0.74}
\definecolor{inputcolor}{rgb}{0.44, 0.035, 0.71}
\definecolor{outputcolor}{rgb}{0.01, 0.61, 0.89}
\usepackage[pagebackref,breaklinks,colorlinks,allcolors=iccvblue]{hyperref}
\usepackage{multirow} 
\usepackage{colortbl}
\usepackage{newtxtext}
\usepackage{newtxmath}
\usepackage{pifont}

\def\paperID{7736} % *** Enter the Paper ID here
\def\confName{ICCV}
\def\confYear{2025}

\title{\textbf{No More Sibling Rivalry: Debiasing Human-Object Interaction Detection}}

\author{
Bin Yang$^{1}$$^*$\hspace{1.5em} Yulin Zhang$^{1}$\thanks{Equal contribution} \hspace{1.5em} Hong-Yu Zhou$^{3}$  \hspace{1.5em} Sibei Yang$^{2}$\thanks{Corresponding author is Sibei Yang.}\\
{$^1$ShanghaiTech University}\hspace{1.5em}
{$^2$School of Computer Science and Engineering, Sun Yat-sen University} \\ {$^3$School of Biomedical Engineering, Tsinghua University} \\
}
\begin{document}
\maketitle
\begin{abstract}

Detection transformers have been applied to human-object interaction (HOI) detection, enhancing the localization and recognition of human-action-object triplets in images. Despite remarkable progress, this study identifies a critical issue—``Toxic Siblings" bias—which hinders the interaction decoder's learning, as numerous similar yet distinct HOI triplets interfere with and even compete against each other both input side and output side to the interaction decoder. 
This bias arises from high confusion among sibling triplets/categories, where increased similarity paradoxically reduces precision, as one’s gain comes at the expense of its toxic sibling’s decline. 
To address this, we propose two novel debiasing learning objectives—``contrastive-then-calibration" and ``merge-then-split"—targeting the input and output perspectives, respectively. 
The former samples sibling-like incorrect HOI triplets and reconstructs them into correct ones, guided by strong positional priors. 
The latter first learns shared features among sibling categories to distinguish them from other groups, then explicitly refines intra-group differentiation to preserve uniqueness. 
Experiments show that we significantly outperform both the baseline (+9.18\% mAP on HICO-Det) and the state-of-the-art (+3.59\% mAP) across various settings.  

\end{abstract}    
\section{Introduction}
\label{sec:intro}

\begin{figure}[t]
\centering
% \vspace{-3mm}
\includegraphics[width=0.46\textwidth]{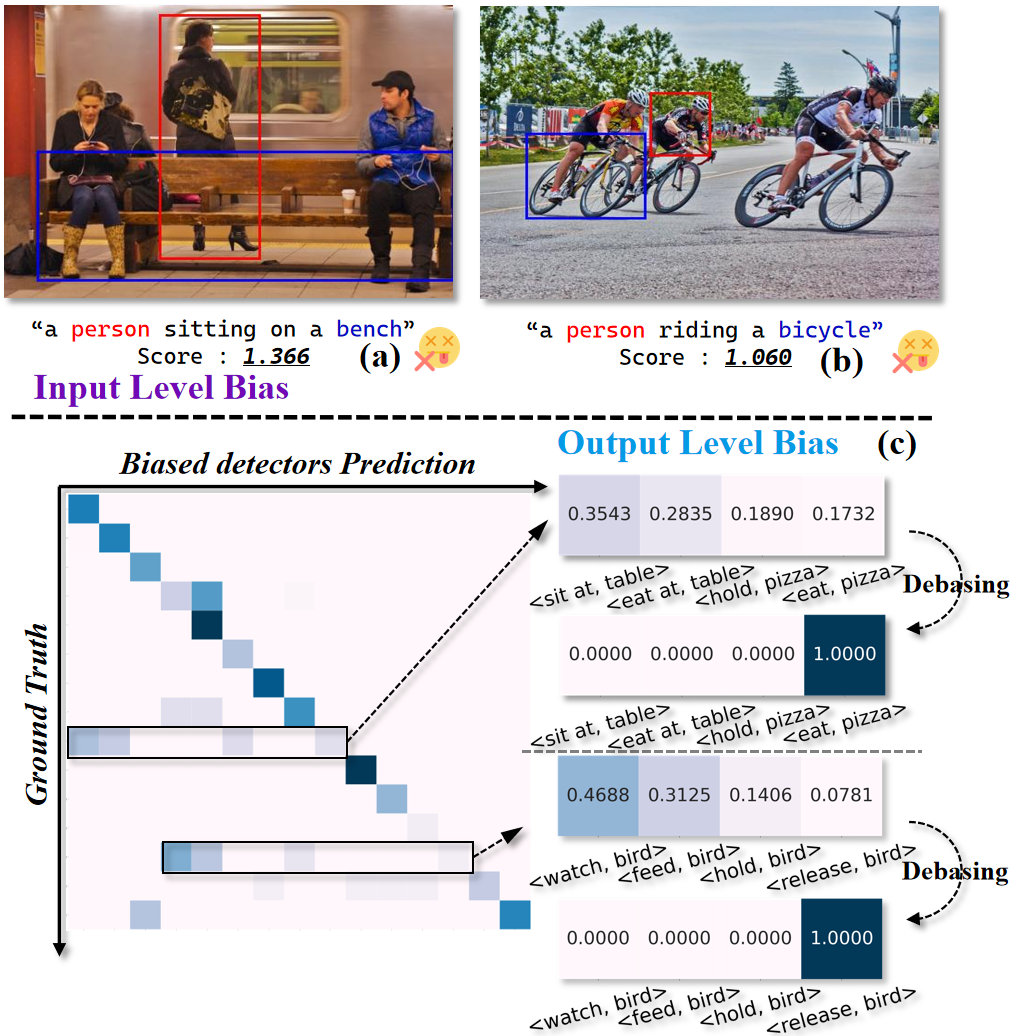}
\caption{\textbf{{Illustration} of ``Toxic Siblings" biases in input and output level} between highly similar HOI triplets and classes. (a)-(b): Input level biases mislead HOI recognition by introducing similar HOIs within images.
For example, in (a), the interaction between the marked person and the stool is misclassified from standing to sitting, influenced by other nearby sibling HOIs humans who are all seated with the stool.
(c) Real confusion matrix in some HOI categories: Output level bias misclassifies rare interactions, such as ``releasing a bird" and ``eating a pizza", by confusing them to sibling common interactions.} 
\vspace{-3mm}
\label{fig:intro1}
\end{figure}

Human-object interaction (HOI) detection~\cite{kim2020uniondet,liao2020ppdm,zhang2022exploring,liao2022gen} aims to detect humans and objects in an image and, importantly, to recognize their interactions as human-action-object triplets. 
Its capability to understand objects and relationships within scenes~\cite{dai2024curriculum, tang2023temporal, lin2021structured, yang2021bottom, he2019non} potentially enables diverse applications in image captioning~\cite{li2023blip,vinyals2015show}, visual question answering~\cite{antol2015vqa,shao2023prompting}, robotics~\cite{diller2024cg}, and human-computer interaction~\cite{preece1994human,hu2024hand}. 
Similar to object detection, HOI detection involves multiple target triplets per image. To handle targets as direct set predictions, recent approaches~\cite{tamura2021qpic,liao2022gen,cao2023detecting} often employ the Detection Transformer (DETR)~\cite{carion2020end, tang2023contrastive, shi2024plain, zhu2025rethinkingquerybasedtransformercontinual} architecture, with sequential detectors: an object detector for identifying humans and objects, followed by an interaction detector for recognizing interactions. 

In turn, compared to object detection~\cite{shi2024plain, shi2024devil, ge2018multi, shi2023edadet}, action-object categories in HOI increase quadratically. Meanwhile, a multitude of distinct human-action-object triplets (HOIs) may share identical human, action, or object components. Consequently, accurately recognizing subtle variations among the vast array of similar HOIs poses a significant challenge, often leading to considerable category confusion.

To enhance the differentiation of HOIs, some approaches~\cite{li2024disentangled,yuanrlip,yuan2023rlipv2} utilize external datasets for explicit data augmentation such as those for object detection~\cite{lin2014microsoft}, scene graphs~\cite{krishna2017visual, chen2024survey}, or action recognition~\cite{carreira2019short} to pretrain object and interaction detectors. 

Others employ external models~\cite{iftekhar2022look,yuanrlip,wang2022learning,qu2022distillation,dong2022category,liao2022gen}, such as VLMs~\cite{ning2023hoiclip,li2023blip}, LLMs~\cite{gpt3, zheng2023ddcot}, and diffusion models~\cite{rombach2022high, ho2020denoisingdiffusionprobabilisticmodels, huang2023free, huang2025mvtokenflow}, to augment data or extract interaction knowledge. 
Alternatively, to enhance interaction recognition and support relational reasoning, recent methods~\cite{xu2019learning,bansal2020detecting,li2020detailed} integrate human-object spatial priors~\cite{bansal2020detecting}, human body pose priors~\cite{li2020detailed}, or fine-grained interaction descriptions alongside object features~\cite{cao2023detecting,luo2024discovering}.

Despite advancements and notable performance improvements, we empirically observe that many HOIs remain difficult to distinguish due to the presence of ``Toxic Siblings Bias", which hinder existing learning mechanisms from effectively helping the detector comprehend the vast and highly similar HOI categories. In the following two paragraphs, we progressively define the ``Toxic Siblings" biases and analyze their impact.

\begin{figure}
    \centering
    \vspace{-3mm}
    \includegraphics[width=0.49\textwidth]{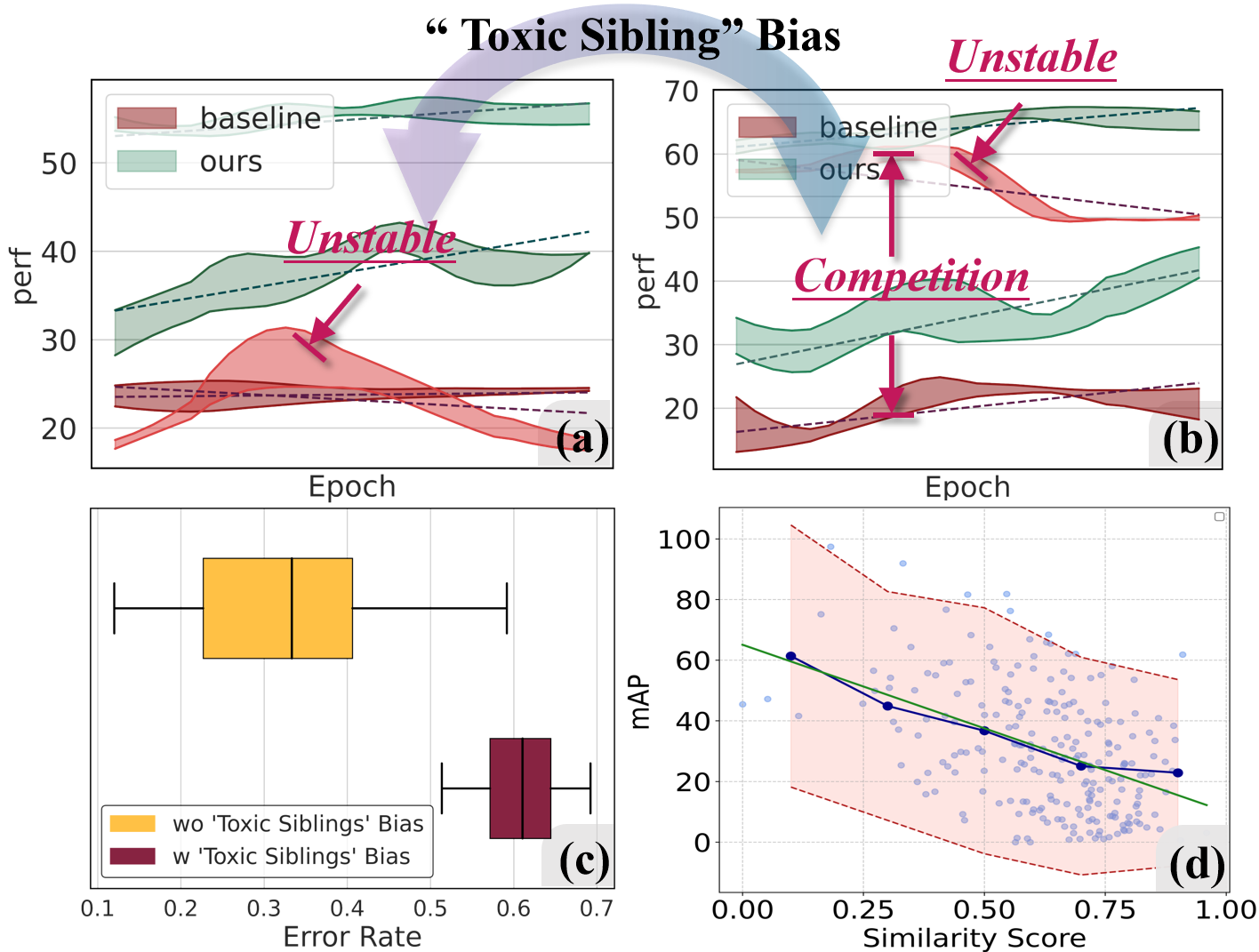}
    \caption{\textbf{{Illustrates} the training patterns and effects of Toxic Siblings Bias in HOI Detection.} \textbf{(a) \& (b)}: The training mAP curves for sibling HOI pairs demonstrate a mutually detrimental learning dynamic (red lines) at both the input-level (a) and output-level (b).
    \textbf{(c)} presents a statistical analysis of the impact of input-level "Toxic Siblings" bias, which significantly increases the error rate. 
    \textbf{(d)} demonstrates that at the output level, the more Toxic Siblings a category has (i.e., the higher the similarity), the worse its performance. 
    (See Appendix A.1 for more details.)}
    \label{fig:intro2}
    \vspace{-5mm}
\end{figure}

 \textbf{What is ``Toxic Siblings Bias"?} 
\textbf{(1) Input bias within single images}: Some HOIs frequently co-occur within the same image, sharing the same object or interaction, as shown in Fig. \ref{fig:intro1}\textcolor{iccvblue}{(a),(b)}. In such cases, the image features input to the interaction decoder contain multiple similar, mutually interfering features for different HOIs, making it harder for the interaction decoder to learn differentiation. 
\textbf{(2) Output bias across different images}: Even when appearing in different images, some categories exhibit semantic similarity—for example, ``watching a bird", ``feeding a bird", ``holdinging a bird", as they all involve ``releasing a bird", as shown in Fig. \ref{fig:intro1}\textcolor{iccvblue}{(c)}. Due to this similarity, the interaction decoder’s output classification heads for sibling categories become highly alike, causing confusion and competition during learning. 

\textbf{How profound is the impact of ``Toxic Sibling" bias on learning?}
Based on the co-occurrence frequency in the same input image and the similarity of class features, we can identify "siblings" for each HOI (details in Appendix \textcolor{red}{A}). The learning curves and statistical results for different groups of siblings reveal their impact on model learning and recognition.
(1) Competition hinders simultaneous accuracy improvement among sibling categories, as illustrated by the two red lines in Fig.~\ref{fig:intro2}\textcolor{iccvblue}{(a)} and Fig.~\ref{fig:intro2}\textcolor{iccvblue}{(b)}. Meanwhile, the severe fluctuations and inverted performance curves indicate that toxic siblings impede further learning.
(2) Then, for the input-level bias, we found that when siblings are present in the input image, the probability of a sample being completely incorrect (mAP=0) increases by an average of \textbf{27.59} (see Fig. \ref{fig:intro2}\textcolor{iccvblue}{(c)}).
(3) Finally, for non-head classes, a 0.1 increase in the average cosine similarity of the classification head initialization (e.g., using CLIP~\cite{radford2021learning}) with other classes results in a \textbf{5.51} drop in AP, as shown in Fig. \ref{fig:intro2}\textcolor{iccvblue}{(d)}.  This indicates that for these HOIs, the more ``Toxic Siblings" they have, the more challenging it becomes for the model to learn to differentiate between them.

In this paper, we aim to mitigate ``Toxic Siblings" biases modifications in the learning mechanism, avoiding additional pretraining data or heavy external models such as LLMs and Diffusion models.
Instead, we introduce only minor modifications to the interaction detector, while \textit{\textbf{emphasizing two novel debiasing objectives—``contrastive-then-calibration" and ``merge-then-split"—to address in-and-out ``Toxic Siblings" biases, respectively}} as shown in Fig~\ref{fig:intro2}. 
The contrastive-then-calibration (C2C) objective samples biased versions of each HOI triplet from its sibling triplets within the same image, then debiases them by realigning and restoring to the correct triplet, guided by strong positional priors. This enables the decoder to resist sibling biases, prioritize spatial cues, and focus on the correct interaction regions for accurate HOI recognition. 

On the other hand, the merge-then-split (M2S) objective, instead of directly requiring the model to distinguish a vast number of HOI categories under the interference of ``Toxic Siblings," first facilitates the rapid learning of shared features among sibling HOIs, followed by explicit differentiation to preserve their uniqueness. Specifically, it leverages the similarity between classes to merge them into superclasses, compensating for insufficient annotations. Then, a discriminative loss is explicitly applied over the top-K similar categories to guide the interaction decoder using class-specific discriminative cues. 

To evaluate the effectiveness of our C2C and M2S learning objectives, we conduct experiments on two standard benchmarks in HOI detection, including HICO-DET~\cite{chao2018learning} and V-COCO~\cite{gupta2015visual}. 
In summary, our contributions are:
\begin{itemize}
\setlength{\itemsep}{0pt}
\setlength{\parsep}{0pt}
\setlength{\parskip}{0pt}
    \item 
    We identify and diagnose a key bias in HOI detection: ``Toxic Siblings," where similar categories interfere with recognition, hindering the learning of numerous HOIs. Statistical analysis shows that "Toxic Siblings" affect model learning at both input and output levels.
    \item 
    We propose the contrastive-then-calibration learning objective to debias input level  ``Toxic Siblings" by realigning and restoring them to the correct ones. 
    \item 
    We propose a novel merge-then-split learning objective to debias output level  ``Toxic Siblings", by first merging to subtly leverage head class data, followed by explicit differentiation to preserve their uniqueness. 
    \item 
    Experiments show that we consistently surpass state-of-the-art methods across all metrics, backbones, and benchmarks, achieving a significant margin of improvement. 
\end{itemize}

\section{Related Work}
\label{sec:related}

\subsection{Human-Object Interaction Detection}

HOI detection aims to identify humans and their associated objects in images, accurately classifying their interaction categories. Existing methods can be broadly categorized into two-stage~\cite{qi2018learning,gupta2019no,wan2019pose,xu2019learning,zhou2019relation,li2019transferable,ulutan2020vsgnet,wang2020contextual,gao2020drg,hou2020visual,li2020hoi,kim2020detecting,li2020detailed,liu2020amplifying,zhou2020cascaded,zhang2021spatially,gkioxari2018detecting,zhou2021cascaded} or one-stage~\cite{kim2020uniondet,liao2020ppdm,wang2020learning,fang2021dirv,tamura2021qpic,liao2022gen,ning2023hoiclip,cao2023detecting,zhang2022exploring} frameworks. 
Two-stage methods first detect humans and objects using an off-the-shelf detector~\cite{carion2020end}, then recognize interactions among each human-object pair by integrating more discriminative features, such as spatial relationships~\cite{xu2019learning,bansal2020detecting,li2020detailed}, structured priors~\cite{zhang2023exploring,zhang2022efficient}, and human poses or keypoints~\cite{gupta2019no,wan2019pose,li2020detailed}. 
In contrast, one-stage methods detect HOI triplets using an end-to-end architecture. Early one-stage methods~\cite{kim2020uniondet,liao2020ppdm,wang2020learning} utilized interaction regions~\cite{kim2020uniondet} or points~\cite{liao2020ppdm} as anchors to provide priors. Inspired by DETR~\cite{carion2020end}, ~\cite{tamura2021qpic} apply DETR to HOI detection by adding additional classification heads. Furthermore, ~\cite{liao2022gen} decouple object detection and relationship prediction through parallel decoders and  ~\cite{chen2021qahoi,ma2023fgahoi} incorporate spatial and semantic priors into queries. 
Unlike existing DETR-like method, we develop novel learning strategies and auxiliary learning objectives, effectively mitigating bias in HOI detection.

\subsection{Data Imbalance in HOI Detection}

The diversity and complexity of human actions exacerbate class imbalance in HOI detection data, where common actions have abundant samples and rare actions are scarce. This imbalance hinders the model's ability to learn effective features for the tail classes.
To address this issue, ~\cite{yang2024open} constructs a balanced dataset using the diffusion model and automated pseudo-labeling. Since pseudo-labels may contain noise~\cite{shi2024part2object, dai2025adaptivelearningfinegrainedgeneralized}, ~\cite{li2024disentangled} decouple HOI detection into object detection and action recognition, leveraging relevant datasets for pre-training. Additionally, ~\cite{yuanrlip,yuan2023rlipv2} transform the classifier to utilize natural language supervision and pre-training on image-text pairs.
Given the robust transferability and generalization capabilities of Visual-Language Models (VLMs) like CLIP~\cite{radford2021learning} across various downstream tasks~\cite{shi2023logoprompt, dai2025freeflyenhancingflexibility}, recent methods~\cite{iftekhar2022look,yuanrlip,wang2022learning,qu2022distillation,dong2022category,liao2022gen} leverage VLMs to help identify tail classes.~\cite{ning2023hoiclip} transfers CLIP's knowledge through queries using a cross-attention mechanism and ~\cite{wu2023end} uses CLIP to obtain action distributions based on cropped human-object pairs. 
Instead of relying on external data or models, we leverage the intrinsic attributes of HOI data to design training strategies that mitigate bias.

\begin{figure*}[t]
\centering
% \vspace{-3mm}
\includegraphics[width=1\textwidth]{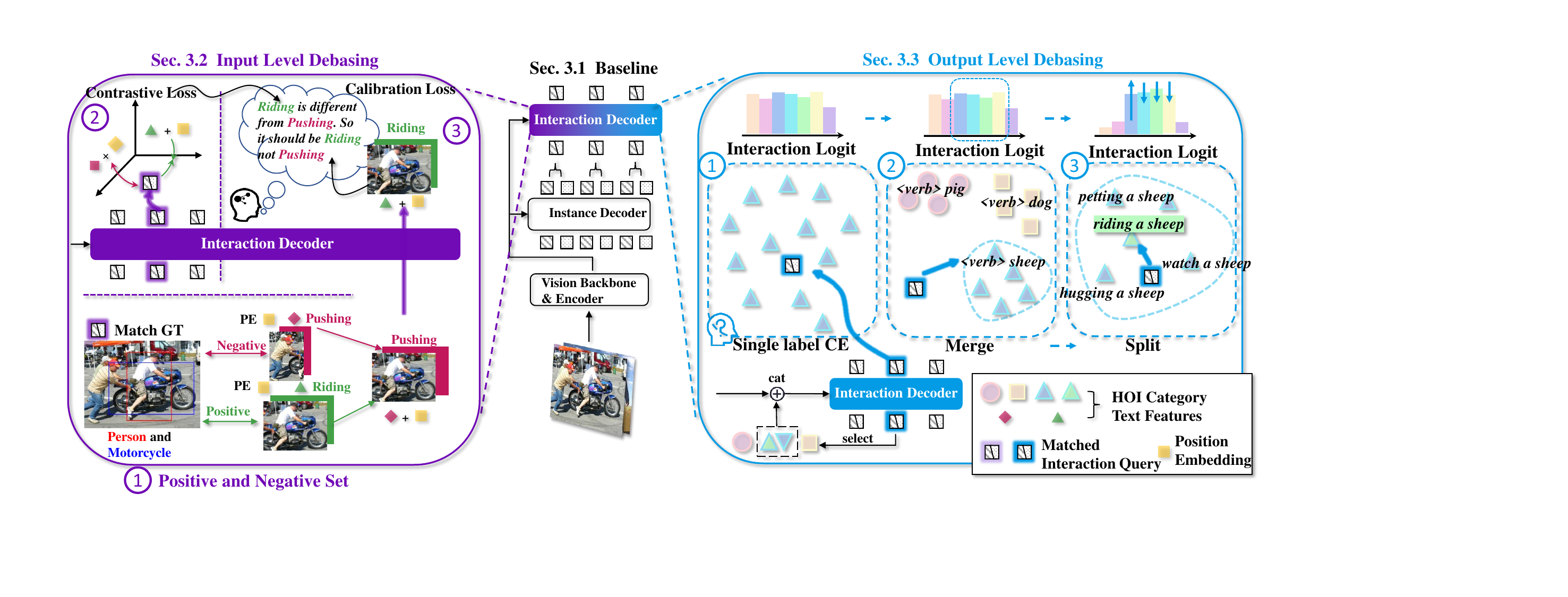}
% \vspace{-7mm}
\caption{\textbf{Overview of overall framework addressing ``Toxic Siblings" biases.} Middle: we use the one-stage two-branch detector~\cite{liao2022gen} as baseline (\cref{method:baseline}), incorporating additional learning objectives in the interaction decoder to address biases. 
\textcolor{inputcolor}{Left (Input “Toxic Siblings"):} 
For each matched query: \textcolor{inputcolor}{\ding{172}} we first identify the positive and negative samples and represent their spatial and semantic features (define in Equ~\ref{equ:set_idenf}); \textcolor{inputcolor}{\ding{173}} (Contrastive) we then enhance their distinctions through contrastive loss (define in \cref{equ:con}). 
\textcolor{inputcolor}{\ding{174}}(Calibration) Building on the ability to differentiate sibling HOIs,  we concatenate  the spatial features of positive samples and the semantic features of negative samples as additional input (define in \cref{equ:cat}), requiring the model to reconstruct them as positive samples (define in \cref{equ:cal}).
% we employ the ``contrastive-then-calibration" learning objective to guide the model’s attention to correct HOI regions and calibrate misclassified interactions. 
\textcolor{outputcolor}{Right (Output ``Toxic Siblings"):} 
% we propose a ``merge-then-split" learning strategy to help the model learn shared features of similar yet distinct HOI categories, while also capturing their differences through class-specific cues.
For each matched query: \textcolor{outputcolor}{\ding{172}} instead of requiring the model to directly distinguish a large number of HOI categories, \textcolor{outputcolor}{\ding{173}} (Merge) we first learn the shared features of sibling HOIs ( \cref{equ:merge}), making them less likely to be confused with other categories, and finally, \textcolor{outputcolor}{\ding{174}} (Split) we preserve their uniqueness by capturing the subtle differences between sibling HOIs (detail in 
\cref{equ:split}).
}

\label{fig:framework}
\vspace{-3mm}
\end{figure*}

\section{Method}
The framework of our proposed model for addressing ``Toxic Siblings" biases is shown in Fig~\ref{fig:framework}. First, we introduce our baseline model~\cite{liao2022gen} (Sec\ref{method:baseline}). Second, to perform input level debiasing, we propose the ``contrastive-then-calibration" (C2C) learning objective (Sec~\ref{method:de_context}). Next, we tackle output level debiasing using the ``merge-then-split" (M2S) approach (Sec~\ref{method:de_class}). Finally, we present the overall loss function of our model (Sec~\ref{method:loss}).
\label{sec:method}
\subsection{Preliminary and HOI Detector}
\label{method:baseline}
\subsubsection{Problem Definition}
Human-object interaction (HOI) detection focuses on understanding the complex relationships between humans and objects. This task can be formalized as detecting a set of \( N \) human-object interaction triplets \( \mathcal{T} = \{ t^\text{hoi}_i \}_{i=1}^{N} \) given an image \( \mathcal{I} \). Each triplet \( t_i^\text{hoi} = (b_i^h, (b_i^o, s_i^o), a_i) \) consists of the following components: \( b_i^h  \) represents the human bounding box, \( (b_i^o, s_i^o)\) denotes the pair of object bounding box and category, and \( a_i \) is the interaction category. Additionally, the pair \( (s_i^o, a_i) \) belongs to the set of all HOI categories \( \mathcal{C} \).

\subsubsection{One-Stage Two-Branch HOI Detector}
Since HOI detection is a set prediction task, following the success of end-to-end DETR in set prediction tasks, several methods (e.g., GEN-VLKT~\cite{liao2022gen}) have introduced DETR~\cite{carion2020end} to HOI detection. They typically include a visual encoder for extracting image features and an instance transformer decoder (\(\operatorname{Decoder_{inst}}\)) for detecting humans and objects, following the traditional DETR architecture. Additionally, they incorporate a serial interaction transformer decoder (\(\operatorname{Decoder_{action}}\)) for detecting interactions.

\noindent \textbf{Visual Encoder} consists of a CNN and a transformer encoder, and is employed to extract image features \( \mathbf{V}^{i} \) from the given image \( \mathcal{I} \).

\noindent \textbf{Instance Decoder} utilizes a set of learnable human-object query pairs to detect humans and objects. Specifically, human and object queries \( \mathbf{Q}^{h}, \mathbf{Q}^{o} \in \mathbb{R}^{\hat{N} \times C} \) are used to query the image features \( \mathbf{V}^{i} \), updating the features and predicting human/object bounding boxes and object categories as \( \{(\hat{b_i^h}, (\hat{b_i^o}, \hat{s_i^o}))\}_{i=1}^{\hat{N}} \), where \( (q_i^h \in \mathbf{Q}^{h}, q_i^o \in \mathbf{Q}^{o}) \) represents a human-object query pair. This process can be formalized as:
\begin{equation}
    (\{(\hat{b_i^h}, (\hat{b_i^o}, \hat{s_i^o}))\}^{\hat{N}}_{i=1}, \mathbf{\tilde{Q}}^h, \mathbf{\tilde{Q}}^o)=\operatorname{Decoder_{inst}}(\mathbf{Q}^h, \mathbf{Q}^o, \mathbf{V}^i)
\end{equation}
where \( \mathbf{\tilde{Q}}^h, \mathbf{\tilde{Q}}^o \in \mathbb{R}^{\hat{N} \times C} \) denote the updated human and object query features, respectively.

\noindent \textbf{Interaction Decoder} follows the Instance Decoder, using its output as interaction queries to identify interaction relationships. Specifically, the updated human and object query features are pooled to initialize the interaction query \( \mathbf{Q}^a = \{ q_i^a \}^{\hat{N}}_{i=1} \in \mathbb{R}^{\hat{N} \times C} \), where $q_i^a=(\tilde{q}^h_i + \tilde{q}^0_i)/{2}$. These interaction query features are then updated using the same mechanism as the Instance Decoder, yielding the updated interaction query features \( \mathbf{\tilde{Q}}^a \), which are used to predict the interaction categories \( \{ \hat{a_i} \}^{\hat{N}}_{i=1} \). This process can be formalized as:
\begin{equation}
    (\{ \hat{a_i} \}^{\hat{N}}_{i=1}, \mathbf{\tilde{Q}}^a ) = \operatorname{Decoder_{action}}(\mathbf{Q}^a, \mathbf{V}^i)
\end{equation}
Following~\cite{cao2023detecting}, we employ an identical, additional interaction decoder to process the visual features from the VLM~\cite{li2023blip, radford2021learning}. For simplicity of notation, we use \(\operatorname{Decoder_{action}}\) to uniformly represent both decoders.

\noindent \textbf{In Detector's Training Stage}, following GEN-VLKT~\cite{liao2022gen}, we initialize the classifier using CLIP~\cite{radford2021learning} text features, which are generated by converting HOI triplets \( \langle \text{Human}, \text{Object}, \text{Verb} \rangle \) into specific text descriptions for different input content (see Appendix for details). This transformation can be denoted as \( \mathrm{F_{text}}(\cdot) \) \label{method:text_feat}, with the label as the input. The Hungarian algorithm~\cite{carion2020end} is used to assign bipartite matching predictions to \( N \) ground-truth labels. Finally, we apply box regression loss \( \mathcal{L}_{b} \), IoU loss \( \mathcal{L}_{u} \), and classification loss \( \mathcal{L}_{c} \) following GEN-VLKT~\cite{liao2022gen}. The total loss function for detector is given by:
\begin{equation}
    \mathcal{L}_{\text{detector}} = \lambda_{b} \sum_{i \in (h, o)} \mathcal{L}_{b}^{i} + \lambda_{u} \sum_{j \in (h, o)} \mathcal{L}_{u}^{j} + \sum_{k \in (o, a)} \lambda_{c}^{k} \mathcal{L}_{c}^{k}.
\end{equation}
\label{method:detecto_loss}

\subsection{\textcolor{inputcolor}{Input ``Toxic Siblings" Debasing}}
\label{method:de_context}
Input ``Toxic Siblings" bias arises when the recognition of a human-object interaction is disproportionately influenced by similar HOIs present in the same image.

To tackle this issue, we empower the decoder to counteract these biases and calibrate misclassified interactions by introducing the ``contrastive-then-calibration" (C2C) debiasing objective. First, we explicitly identify instances within an image that are siblings yet represent different HOI triplets, and then perform contrastive learning between these instances to learn their distinctions. 
Finally, through auxiliary training guided by strong spatial priors, the decoder further calibrates misclassified HOI triplets by reconstructing them to the correct ones.

\noindent \textbf{Positive and Negative Set.} For an HOI triplet, we define its positive set as the set of triplets within the same image that share the same interaction-object label, while its negative set consists of other triplets that share either the interaction or object label. Specifically, for a triplet \( t^\text{hoi}_i = (b_i^h, (b_i^o, s_i^o), a_i)) \), its positive set \( \mathcal{P}_i \) and its negative set \( \mathcal{N}_i \) can be represented as {follows}:

\begin{equation}
    \label{equ:set_idenf}
    \begin{aligned}
        \mathcal{P}_i & = \{t^\text{hoi} \in \mathcal{T} \mid s^o = s_i^o \land a = a_i\}, \\
        \mathcal{N}_i & = \{t^\text{hoi} \in \mathcal{T} \mid (s^o = s_i^o \land a \neq a_i) \lor (s^o \neq s_i^o \land a = a_i)\}.
    \end{aligned}
\end{equation}
During training, for a query \(\hat{q}_i^a\), it shares the same positive and negative set as its matched ground-truth triplet.

\noindent \textbf{Triplet Feature Representation.} To comprehensively represent the semantic and spatial information of a triplet, we obtain its feature representation by concatenating semantic features with the spatial features of the human and object. Specifically, for a query, we use its updated query feature \(\hat{q}_i^a\) as the semantic feature, while for a ground-truth triplet \(t^\text{hoi}_i\) without a query feature, we use the text feature generated by the transformation \(\mathrm{F_{text}}(\cdot)\) instead:
\begin{equation}
    \label{equ:feat_rep}
   \hat{v}^{\text{tri}}_i = \operatorname{cat}(\hat{q}^a_i, \hat{f}^h_i, \hat{f}^o_i), \quad
{v}^{\text{tri}}_i = \operatorname{cat}(\mathrm{F_{text}}(s^o_i, a_i), {f}^h_i, {f}^o_i), 
\end{equation}
where \(\hat{v}^{\text{tri}}_i\) and \({v}^{\text{tri}}_i\) represent the triplet features for the query \(\hat{q}_i^a\) and the ground-truth triplet \(t^\text{hoi}_i\), respectively. Here, \(\hat{f}^h_i, \hat{f}^o_i, {f}^h_i,\) and \({f}^o_i\) denote the spatial features of the human and object, generated by applying a linear layer to the corresponding bounding box inputs.

\noindent \textbf{Input Debias Contrastive Loss.} For each triplet, in order to increase its distinction from the negative set and reduce the distance to the positive set, we apply a contrastive loss. Specifically, the loss~\cite{oord2018representation} is defined as:
\begin{equation}
    \label{equ:con}
    \mathcal{L}_\text{{con}}=-\frac{1}{N} \sum_{i=1}^{N} \log \frac{
\sum_{{v}^\text{tri}_j\in \mathcal{P}_i} \exp \left(\hat{v}_{i}^\text{tri} \odot{v}^\text{tri}_j / \tau\right)}{\sum_{{v}^\text{tri}_k\in \mathcal{E}_i} \exp \left(\hat{v}_{i}^\text{tri} \odot{v}^\text{tri}_k / \tau\right)} {,}
\end{equation}
where \(\tau\) is a temperature parameter for scaling the contrastive logits, and \(\hat{v}_{i}^\text{tri}\) and \({v}^\text{tri}_j\) are defined in \cref{equ:feat_rep}.

\noindent \textbf{Input Debias Calibration Loss.} Through contrastive loss, the features of different HOI triplets that are siblings become more distinguishable. Building on this, we aim to enhance the interaction decoder's ability to focus on the correct interaction regions based on spatial information, enabling it to identify and correct misclassified triplets. Therefore, during training, we introduce queries with incorrect semantic features and enforce their features to be calibrated and reconstructed correctly. 

Specifically, for an HOI triplet feature \(v^{\text{tri}}_i\) from the positive set, we randomly replace its semantic feature with one sampled from the negative set \(\mathcal{N}_i\) and concatenate it with the correct spatial features of the human ${f}^h_i$ and object ${f}^o_i$, thereby providing a strong spatial prior:
\begin{equation}
    \label{equ:cat}
    \begin{aligned}
    \underline{v}^\text{tri}_i &= \operatorname{cat}(\mathrm{F_{text}}(s^o,a),{f}^h_i,{f}^o_i),
    \end{aligned}
\end{equation}
where $(s^o,a) = \operatorname{sample}(\mathcal{N}_i)$. This replaced feature $\underline{v}^\text{tri}_i$ is then used as an additional query input to the interaction decoder, and we require it to be corrected back to its original feature:
\begin{equation}
    \label{equ:cal}
    \begin{aligned}
   (\{\hat{a_i} \}^{\hat{N}}_{i=1},  \mathbf{\tilde{Q}}^a, \mathbf{\tilde{Q}}^{a}_\text{add} ) &=\operatorname{Decoder}_{\text{action}}(\mathbf{Q}^a, \mathbf{V}^i,\mathbf{Q}^{a}_\text{add}),  \\
    \mathcal{L}_{\text{cal}} &= - \frac{1}{N} \sum_{i=1}^{N} \left\| \overline{v}^{tri}_i - v^{tri}_i \right\|_1 ,
\end{aligned}
\end{equation}
where \(\mathbf{Q}^{a}_\text{add} = \{\underline{v}^\text{tri}\}_{i=1}^{N}\) is an additional query set with semantic labels replaced, and \(\mathbf{\tilde{Q}}^{a}_\text{add} = \{\overline{v}^\text{tri}_i\}_{i=1}^{N}\) is the correction result of the interaction decoder for \(\mathbf{Q}^{a}_\text{add}\). We use the L1 loss, denoted as \(\mathcal{L}_{\text{cal}}\), to evaluate the effectiveness of the correction. 
Notably, we use an attention mask to ensure that the additional queries do not interfere with the original queries \(\mathbf{Q}^a\).

\subsection{\textcolor{outputcolor}{Output ``Toxic Siblings" Debiasing}}
\label{method:de_class}

In HOI detection, where the number of categories is several times larger than in object detection or classification tasks, directly forcing the model to learn the uniqueness of each category is prone to "Toxic Siblings" bias due to their high semantic similarity.

To address this, we propose leveraging the shared features among sibling HOIs (e.g., the head class "watch a bird" and the tail class "release a bird", as shown in Fig.~\ref{fig:intro1} \textcolor{iccvblue}{(c)}) to distinguish them from a vast number of other categories, which we refer to as the Merge Learning Objective. Building on this, to further capture the subtle differences within sibling categories, we introduce the Split Learning Objective.

\subsubsection{Merge Learning Objective}
\label{method:merge}
To better learn the general features shared between siblings, we utilize clustering to obtain superclasses for all HOI categories. The interaction decoder is then required to perform superclass classification.

Specifically, by applying the transformation $\mathrm{F_{text}}(\cdot)$, we generate $N_1$ interaction and $N_2$ object text features for the corresponding interaction and object categories. Then, we perform clustering on the $N_1$ interaction and $N_2$ object text features separately, resulting in $M_1$ super interaction categories and $M_2$ super object categories. They can be combined into \(M\) super interaction-object categories, where \(M = M_1 \times M_2\). Through the hierarchical relationships in the clustering, we assign a superclass label to each interaction-object category. Therefore, the label set $\{(b_i^h, (b_i^o, s_i^o), a_i)\}^{n_{gt}}_{i=1}$ can be extended to $\{(b_i^h, (b_i^o, s_i^o), a_i, {m_i})\}^{n_{gt}}_{i=1}$, where ${m_i}$ denotes the superclass label.

Finally, we incorporate an additional classification output into the \(\operatorname{Decoder_{action}}\) and apply the classification loss, denoted as \(\mathcal{L}_\text{merge}\):
\begin{equation}
    \label{equ:merge}
        \begin{aligned}
    &(\{\hat{a}_i\}_{i = 1}^{\hat{N}}, \mathbf{\tilde{Q}^a}, {\{\hat{m}_i\}_{i = 1}^{\hat{N}}}) = \operatorname{Decoder_{action}}(\mathbf{Q^a}, \mathbf{V}^i), \\
    &\mathcal{L}_\text{merge} = \operatorname{CrossEntropy}(\{m_i\}_{i = 1}^{{N}}, \{\hat{m}_i\}_{i = 1}^{{N}}).
    \end{aligned}
\end{equation}
Here, $\hat{m}_i$ represents the predicted superclass for the $i$-th interaction query, and $\{{m}_i\}_{i=1}^{\hat{N}}$ represents the predicted results for the ground-truth matching queries.

\subsubsection{Split Learning Objective}
\label{method:split}
By employing hierarchical classification based on HOI categories and superclasses, the model not only learns fine-grained category information but also learns general features shared among different categories. However, one potential concern is that this approach may make it more difficult to distinguish between HOI categories that share similar features. To address this, we propose the split learning objective, which aims to help the model better differentiate between these categories by identifying the most similar $k$ HOI categories for each query and requiring the model to classify the query correctly among them.

Specifically, for query $q_i^a$, we first identify its $k_1$ most similar categories, $\{c_j\}_{j=1}^{k_1}$, based on the similarity between $q_i^a$ and all HOI categories, computed as follows:
\begin{equation}
    (\{c_j\}_{j=1}^{k_1}) = \operatorname{topk} \ \operatorname{sim}(q_i^a, T),
\end{equation}
where $T$ denotes the set of all category features, which are extracted using the transformation $\mathrm{F_{text}}(\cdot)$.

Next, we augment the image features with category features to assist interaction queries in distinguishing between the most similar $k_1$ categories, guided by class-specific information and split learning objective. Instead of concatenating all top-$k_1$ category features to the image features for every query, we select and concatenate only the $k_2$ most frequent category features among the top-$k_1$ categories across all queries. This approach not only eliminates noise from background queries but also reduces computational cost. Specifically, for the $k_2$ most frequent category features, $T^{k_2} = \{t_i\}_{i=1}^{k_2}$, and their corresponding categories $\{c_j\}_{j=1}^{k_2}$, the interaction decoder's process can be rewritten as:
\begin{equation}
    (\{a_i\}_{i=1}^{\hat{N}}, \mathbf{\tilde{Q}^a}, {\{\hat{m}_i\}_{i=1}^{\hat{N}}}) = \operatorname{Decoder_{action}}(\mathbf{Q^a}, {\operatorname{cat}(\mathbf{V}^i, T^{k_2})}).
\end{equation}

To ensure consistency between the input category features and the learning objectives, the split learning objective is only applied to interaction features that have been assigned a label, where the label belongs to the set $\{c_j\}_{j=1}^{k_2}$. These interaction features are denoted as $\{\tilde{q}^a_i\}^{N_\text{split}}_{i=1} \subseteq \{\tilde{q}^a_i\}^{N}_{i=1}$, with $N_\text{split}$ being the number of interaction features that satisfy the aforementioned condition. The corresponding text features are $\{t_i\}^{N_\text{split}}_{i=1}$. Therefore, the split learning loss $\mathcal{L}_{\text{split}}$, can be written as:
\begin{equation}
    \label{equ:split}
    \mathcal{L}_{\text{split}} = -\frac{1}{N_\text{split}} \sum_{i=1}^{N_\text{split}} \log \frac{
 \exp \left(q^a_i \odot t_i / \tau\right)}{\sum_{j=1}^{k_2}\exp \left(q^a_i \odot t_j / \tau\right)},
\end{equation}
where $\tau$ is a temperature parameter used to scale the contrastive logits.

\subsection{Overall Learning Objective}
\label{method:loss}
Combining Sections ~\ref{method:de_context} and ~\ref{method:de_class}, the total loss is:
\begin{equation}
    \mathcal{L}_\text{all} = \lambda_{1}\mathcal{L}_{\text{detector}} + \lambda_{2}\mathcal{L}_{\text{con}} + \lambda_{3}\mathcal{L}_{\text{cal}} + \lambda_{4}\mathcal{L}_{\text{merge}} + \lambda_{5}\mathcal{L}_{\text{split}},
\end{equation}
where \( \lambda_{1}, \lambda_{2}, \lambda_{3}, \lambda_{4}, \lambda_{5} \) are the weights for each loss.
\section{Experiment}
\label{sec:method}

\begin{table*}[t]
  \centering
  \small
   \captionsetup{font=small}
  % \caption{\small{{Quantitative results} on HICO-DET\!~\cite{chao2018learning} .}}
  % \label{}
  \setlength\tabcolsep{4pt}
  \renewcommand\arraystretch{1.1}
  \resizebox{0.8\textwidth}{!}{
        \begin{tabular}{l l l c c c c c c c c c}
       \toprule[1.0pt]
       & & & & \multicolumn{6}{c}{HICO-DET}  & \multicolumn{2}{c}{V-COCO}\\
      & \multirow{-1.5}{*}{Back}& & \multirow{-1.5}{*}{External}& \multicolumn{3}{c}{Default}& \multicolumn{3}{c}{Known Object} \\
      \multirow{-3}{*}{Method} & \multirow{-1.5}{*}{bone}&\multirow{-3}{*}{VLM}& \multirow{-1.5}{*}{Models} & Full& Rare& Non-Rare& Full& Rare& Non-Rare  & \multirow{-2}{*}{$AP^{S1}_{role}$} & \multirow{-2}{*}{$AP^{S2}_{role}$}\\
      \midrule
      DiffHOI~\cite{yang2023boosting} &R50 & & SD &34.41 &31.07 &35.40 &37.31 &34.56 &38.14 &61.1 &63.5\\
      PViC~\cite{zhang2023exploring}  &R50 & & &34.69 &32.14 &35.45 &38.14 &35.38 &38.97 &62.8 &67.8\\
      % EP-HOI~\cite{wu2024exploring} &R50 & & & 35.86 & 32.48 & 36.86 & 39.48 & 36.10 & 40.49 \\
      GEN-VLKT\cite{liao2022gen}& R50& CLIP & &33.75& 29.25& 35.10& 37.80& 34.76& 38.71& 62.4& 64.4\\
      HOICLIP~\cite{ning2023hoiclip}  &R50 &CLIP & & 34.69 &31.12 &35.74 &37.61 &34.47 &38.54 &63.5 &64.8\\
      ViPLO~\cite{park2023viplo} & R50&CLIP && 34.95 &33.83 &35.28 &38.15 &36.77 &38.56 &60.9& 66.6\\
      ADA-CM~\cite{lei2023efficient} & R50 & CLIP && 38.40 &37.52 &38.66 &- &- &- &58.6 &64.0\\
      BCOM~\cite{wang2024bilateral}  & R50 & CLIP &&39.34 &39.90 &39.17 &42.24 &42.86 &42.05 &65.8 &69.9\\
     \cellcolor{violet!7}\textbf{Ours}  & \cellcolor{violet!7}R50 & \cellcolor{violet!7}CLIP &\cellcolor{violet!7}& \cellcolor{violet!7}\textbf{42.93} & \cellcolor{violet!7}\textbf{42.41} & \cellcolor{violet!7}\textbf{43.11} & \cellcolor{violet!7}\textbf{44.97} & \cellcolor{violet!7}\textbf{44.20} & \cellcolor{violet!7}\textbf{45.23}  &\cellcolor{violet!7}\textbf{69.8} &\cellcolor{violet!7}\textbf{72.1}\\
         \midrule
      UniHOI~\cite{cao2023detecting} &R50& BLIP2 & LLM& 40.06 &39.91 &40.11 &42.20 &42.60 &42.08& 65.6 & 68.3 \\

      SICHOI~\cite{luo2024discovering}  &R50& BLIP2 &LLM+SAM& 41.79  &42.38 &41.61 &44.27 &43.64 &44.46&67.9 &72.8\\

       \cellcolor{violet!7}\textbf{Ours} & \cellcolor{violet!7}R50 & \cellcolor{violet!7}BLIP2 &\cellcolor{violet!7}& \cellcolor{violet!7}\textbf{43.98} &\cellcolor{violet!7}\textbf{43.27} &\cellcolor{violet!7}\textbf{44.00} &\cellcolor{violet!7}\textbf{45.90} &\cellcolor{violet!7}\textbf{45.29} &\cellcolor{violet!7}\textbf{46.32} & \cellcolor{violet!7}\textbf{71.1}& \cellcolor{violet!7}\textbf{73.3}\\
      \midrule
      GEN-VLKT\cite{liao2022gen}& R101& CLIP  &&34.95& 31.18& 36.08& 38.22& 34.36& 39.37& 63.6& 65.9\\
      ViPLO~\cite{park2023viplo} & R101&CLIP && 37.22 &35.45 &37.75 &40.61 &38.82 &41.15 &62.2& 68.0 \\
      CyCleHOI~\cite{wang2024cyclehoi}& R101&CLIP & SD & 37.79 &34.22 &38.61 &41.13 &37.43 &42.06 &66.5& 68.5 \\
      \cellcolor{violet!7}\textbf{Ours} & \cellcolor{violet!7}R101 & \cellcolor{violet!7}CLIP &\cellcolor{violet!7}& \cellcolor{violet!7}\textbf{43.82} &\cellcolor{violet!7}\textbf{43.11} &\cellcolor{violet!7}\textbf{44.10} &\cellcolor{violet!7}\textbf{45.84} &\cellcolor{violet!7}\textbf{45.25} &\cellcolor{violet!7}\textbf{46.17}  &\cellcolor{violet!7}\textbf{70.1} &\cellcolor{violet!7}\textbf{72.4}\\
      
      \bottomrule[1.0pt]
      \end{tabular}
    }
  \caption{Comparison with state-of-the-art models on HICO-DET\!~\cite{chao2018learning} dataset in Default and Known Object settings and V-COCO\!~\cite{gupta2015visual} dataset in $AP^{S1}_{role}$ and $AP^{S2}_{role}$. SD represents Stable Diffusion~\cite{rombach2022high}, LLM represents GPT3~\cite{gpt3}, SAM represents Segment Anything~\cite{kirillov2023segment}.}
\vspace{-12pt}
  \label{main_hico}
\end{table*}

\subsection{Experimental Setups}
\noindent \textbf{Datasets.} We train and evaluate on two datasets: HICO-DET~\cite{chao2018learning} and V-COCO~\cite{gupta2015visual}. HICO-DET comprises 47,776 images, with 80 object categories and 117 verb categories, resulting in 600 human-object interaction (HOI) categories. It is split into three subsets: (i) Full, covering all 600 HOI categories; (ii) Rare, containing 138 categories with fewer than 10 samples; and (iii) Non-Rare, with the remaining 462 categories. V-COCO, a subset of MS-COCO~\cite{lin2014microsoft}, includes 10,346 images, 24 interaction types, and 80 object categories. We follow evaluation protocols of GEN-VLKT~\cite{liao2022gen} and tested performance on both datasets.

\noindent \textbf{Implementation details.}\label{im_detail} We adopt the GEN-VLKT~\cite{liao2022gen} method, using the DETR~\cite{carion2020end} architecture which is initialized with pre-trained parameters from MS-COCO~\cite{lin2014microsoft}. For fair comparison, we use either the CLIP~\cite{radford2021learning} or BLIP2~\cite{li2023blip} model as the visual-language (VL) foundation model. Similar to GEN-VLKT, we employ the AdamW~\cite{loshchilov2017decoupled} optimizer with a weight decay of \(10^{-4}\). The model is trained for 90 epochs with an initial learning rate of \(10^{-4}\), reduced by a factor of ten at the 60th epoch. Hyperparameters are set as follow the configuration in~\cite{liao2022gen}: \( \lambda_{b} = 2.5 \), \( \lambda_{u} = 1 \), and \( \lambda_{c}^{k} = 1 \), with additional settings for loss weight \(\lambda_{1} = 1\), \(\lambda_{2} = 1\), \(\lambda_{3} = 0.5\), \(\lambda_{4} = 1\), \(\lambda_{5} = 0.5\), and the hyper-parameter mention in \ref{method:split} and \ref{method:merge}: \(k_1 = 2, k_2 = 10\), \(M = 25\) . All experiments are conducted on 8 Nvidia A40 GPUs with a batch size of 8. 

\subsection{Comparison with State-of-the-Arts}

For both HICO-DET and V-COCO, as shown in Table \ref{main_hico}, our method achieves significant average improvements of 3.66\% and 3.69\%, respectively, over existing SOTA methods across various datasets, backbones, and VLM configurations.
Specifically, 

(1) compared to the SOTA method using CLIP as the VLM, our approach achieves at least 2.72\% gains on HICO-DET. Similarly, for the V-COCO dataset, our method achieves a minimum improvement of 2.2\%. These results demonstrate that our method provides superior HOI detection capabilities across identical backbones and VLMs, regardless of dataset and partition.

(2) Compared to our baseline, GEN-VLKT~\cite{liao2022gen}, we achieve at least 7.17\% and 6.5\% improvements on the HICO-DET and V-COCO datasets, respectively. Notably, our method shows a substantial gain of 13.16\% on rare categories, reaching 42.41\%, attributable to the ``merge-then-split" learning objective, which effectively output level ``Toxic Siblings" bias.

(3) On the same datasets and partitions and using CLIP as the VLM, our method with only an R50 backbone outperforms the SOTA method CyCleHOI~\cite{wang2024cyclehoi} with R101 by an average of 4.49\%. This result indicates the parameter efficiency of our proposed method.

(4) Furthermore, our method is adaptable across different VLMs, delivering a consistent average improvement of 3.48\% whether using CLIP or BLIP2. This indicates that our approach is transferable across multiple VLM setups. 

(5) As ~\cite{yuan2023rlipv2} mentioned, HOI task suffers from the lack of annotated data, which effect some methods~\cite{cao2023detecting,luo2024discovering,wang2024cyclehoi,yang2023boosting} to incorporate heavy external models, such as Stable Diffusion~\cite{rombach2022high}, LLM~\cite{gpt3}, and SAM~\cite{kirillov2023segment}, to potentially augment the data. Notably, our method even outperforms works~\cite{cao2023detecting,luo2024discovering,wang2024cyclehoi,yang2023boosting} with these heavy external models.

(6) Table~\ref{hico_zero_shot} shows the results of our method across four zero-shot settings, With the R50 backbone on HICO-DET dataset. Ours method demonstrates competitive performance in HOI recognition compared to supervised methods in Table ~\ref{main_hico}. This highlights the widespread existence of the toxic siblings bias, which we effectively mitigate.

\begin{table}[h]
  \centering
  \small
   \captionsetup{font=small}

  \setlength\tabcolsep{4pt}
  \renewcommand\arraystretch{1.1}
  \resizebox{0.48\textwidth}{!}{
        \begin{tabular}{l c c c c c c c}
       \toprule[1.0pt]
      Method & Backbone& External Models & & UV & UO&NF& RF \\
      \midrule

       \midrule
      GEN-VLKT &R50& CLIP & useen&20.96 & 10.51 &25.05 &21.36  \\

        UniHOI &R50& BLIP2+LLM& useen&26.05 & 19.72 &28.45 &28.68 \\
        
      SICHOI  &R50& BLIP2+LLM+SAM & useen&- & -  &34.52 &34.24 \\

       \cellcolor{violet!7}\textbf{Ours} & \cellcolor{violet!7}R50 & \cellcolor{violet!7}CLIP  &\cellcolor{violet!7} useen& \cellcolor{violet!7}\textbf{30.47} & \cellcolor{violet!7}\textbf{31.01} &\cellcolor{violet!7}\textbf{36.12} &\cellcolor{violet!7}\textbf{35.48} \\
      
      \midrule

      \bottomrule[1.0pt]
      \end{tabular}
    }
  \caption{Zero-Shot Results on HICO-DET.}
\vspace{-12pt}
  \label{hico_zero_shot}
\end{table}

\subsection{Ablation Study}

In this section, we conduct a series of experiments to analyze the effectiveness of our learning objectives. All experiments are conducted in HICO-DET dataset and using GEN-VLKT-s as baseline, as shown in Table \ref{ab_com}. 

\noindent  \textbf{Ablation on Input Level Debasing.}
To validate the effectiveness of our proposed input debasing strategy, we conduct ablation experiments based on the baseline model. 
%(+HOR Mask) 

(1) First, we improve the baseline using the human-object mask attention. (see \textcolor{iccvblue}{A.3} for implementation details and Tab. \textcolor{iccvblue}{A3} for result without HOR mask)

(2) However, relying solely on mask attention is insufficient to eliminate interference from adjacent ``sibling". %, as shown in Fig.~\ref{fig:intro1} (a). 
Therefore, we introduce a contrastive loss to enhance the model's discrimination between different HOI triplets, helping it better distinguish interactions within a single image. This reduction in context bias further improved performance, reaching 36.82\% (+Contrastive Loss).
%(+Correction Loss) 

(3) Finally, we introduce a calibration loss in the training of the interaction decoder to better correct misclassified samples caused by  bias. As shown in Table \ref{ab_com}, line 4, this addition led to a further performance increase of 1.21\% (+Calibration Loss), demonstrating that calibration loss effectively mitigates the impact of bias, thus enhancing HOI detection capabilities.

\begin{figure}[t]
    % \vspace{-3mm}
    \centering
    \includegraphics[width=0.95\linewidth]{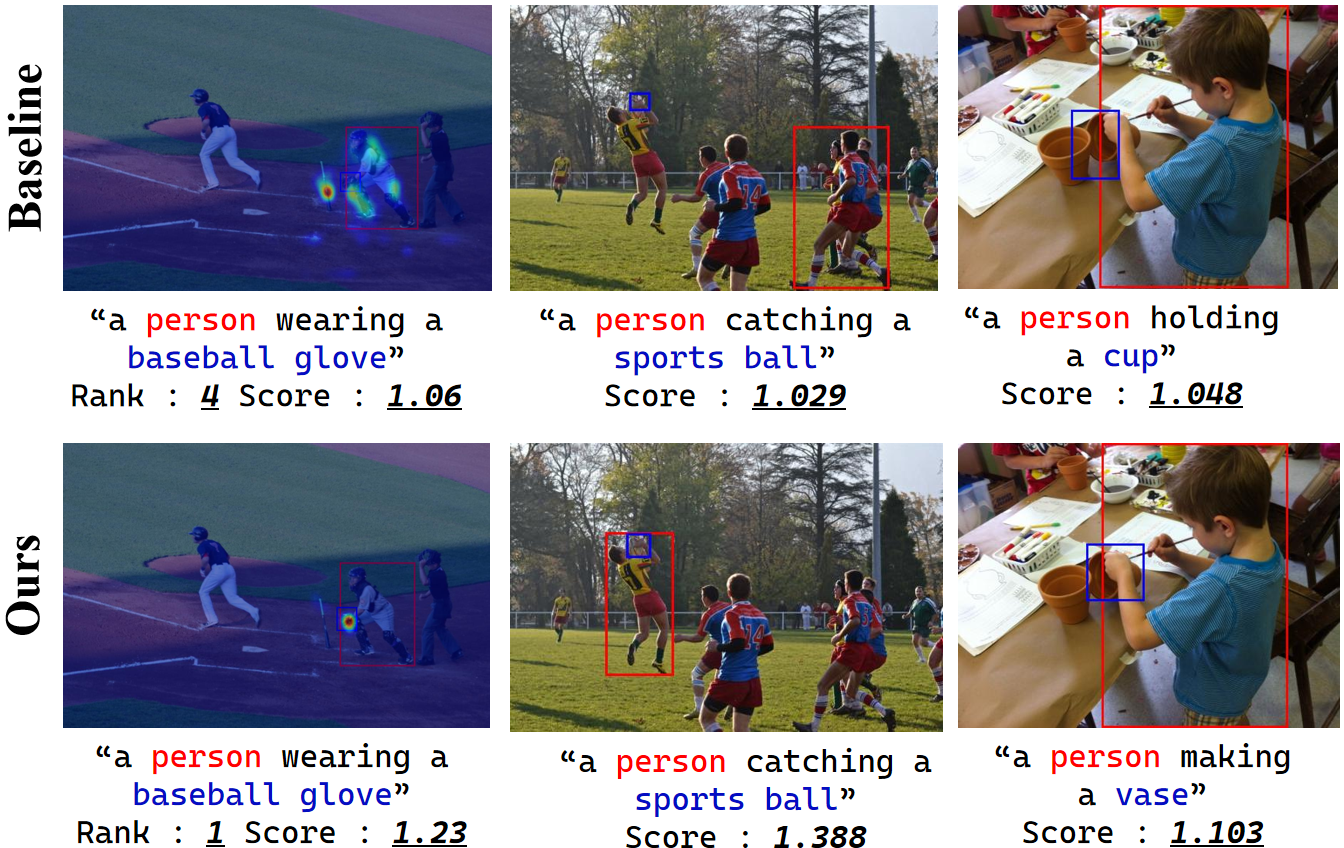}
    \vspace{-2.5mm}
    % \caption{}
            % \vspace{-0.1cm}
    \caption{Visualization of our prediction results and comparison to GEN-VLKT~\cite{liao2022gen}.}
    \label{fig:vis}
    \vspace{-5mm}
\end{figure}

\begin{figure}[t]
    % \vspace{-3mm}
    \centering
    \includegraphics[width=0.95\linewidth]{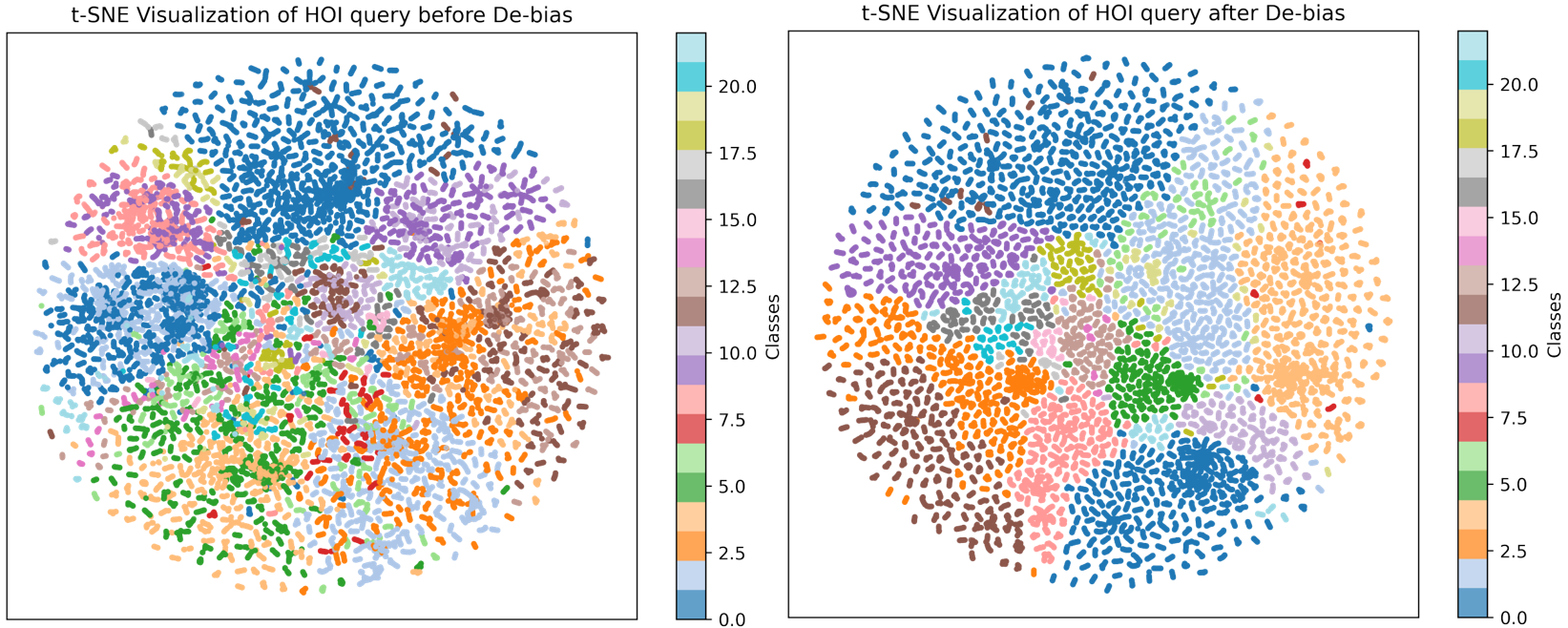}
    \vspace{-2.5mm}
    % \caption{}
            % \vspace{-0.1cm}
    \caption{Visualization of our prediction results and comparison to GEN-VLKT~\cite{liao2022gen}.}
    \label{fig:vis_tsne}
    \vspace{-5mm}
\end{figure}

\begin{table}[h]
\centering
\small
  
\resizebox{0.4\textwidth}{!}{
\begin{tabular}{lccc}
\toprule
\multirow{2.2}{*}{Method} & \multicolumn{3}{c}{\textbf{HICO-DET (Default)}} \\
\cline{2-4} 
 & Full & Rare & Non-Rare \\
\hline
\midrule
Baseline(GEN-VLKT-s)               & 33.75         & 29.25        & 35.10                   \\
\midrule
Input Level Debasing:\\

+HOR mask                     & 35.37         & 30.70         & 36.51                 \\
+Contrastive Loss                    & 36.82         & 32.29         & 37.92                   \\
+Calibration Loss                      & 38.03         & 34.05         & 39.80                    \\
\midrule
Output Level Debasing: \\
+Merge Loss             & 39.88         & 36.12         & 40.20                    \\
+Split Loss              & \textbf{42.93} & \textbf{42.41} & \textbf{43.11}    \\
\bottomrule
\end{tabular}
}
\caption{Ablation study on HICO-DET\!~\cite{chao2018learning} with Default setting.}
\vspace{-3mm}
\label{ab_com}
\end{table}

\noindent \textbf{Ablation on Output Level Bias.}
To validate the effectiveness of our ``merge-then-split" learning objective in mitigating output ``Toxic siblings" bias, we conduct incremental experiments based on Table \ref{ab_com}, line 5, gradually adding the merge and split learning objectives. 

(1) First, we introduce the merge learning objective to the interaction decoder(+Merge Learning Objective). By incorporating a hierarchical loss at the superclass level, this approach effectively leverages shared features to distinguish between a vast number of other HOIs, resulting in a performance gain of 1.85\%.

(2) To refine class distinctions, ensuring the model effectively captures inter-class differences and enhances fine-grained interaction understanding, we introduce the split learning objective(+Split Learning Objective). This learning objective boosts performance to 42.93\%, indicating that the split objective enhances the model's ability to distinguish among similar classes, alleviating concerns about potential difficulties in differentiating certain classes. 

(3) This synergistic ``merge-then-split" strategy elevates performance on the HICO-DET dataset from 38.03\% to 42.93\%, significantly surpassing the baseline and demonstrating its effectiveness in addressing output level bias.

\begin{table}[h]
\centering
 \small
\begin{tabular}{lccc}
\toprule
 Ablation on & \multicolumn{3}{c}{\textbf{HICO-DET (Default)}}  \\
\cline{2-4} 
 \# of $k_1$ and $k_2$ & Full & Rare & Non-Rare  \\
\midrule
$k_1=2,k_2=5$ & 41.20 & 41.05 & 42.19 \\
$k_1=2,k_2=10$ & \textbf{42.93} & \textbf{42.41} & \textbf{43.11} \\
$k_1=2,k_2=15$ & 42.72 & 42.35 & 42.97 \\
\midrule
\# of cluster & & & \\
\midrule
$M_1 \times M_2 = 3 \times 3$ & 41.88 & 41.37 & 42.05 \\
$M_1 \times M_2 = 5 \times 5$ & \textbf{42.93} & \textbf{42.41} & \textbf{43.11} \\
$M_1 \times M_2 = 7 \times 7$ & 42.15 & 41.40 & 42.57 \\
\bottomrule
\end{tabular}
% \vspace{-3mm}
\caption{Ablation study on Hyper-parameters.}
\vspace{-3mm}
\label{ab_topK}
\end{table}

\noindent \textbf{Ablation on Hyper-parameters.} 

We conduct ablation on hyper-parameter $k_1$ and $k_2$ in Sec \ref{method:split} and $M_1$ and $M_2$ in Sec \ref{method:merge}.  $k_1$ and $k_2$ represent the range of category features selection in split, while $M_1$ and $M_2$ respectively denote the number of action and object cluster centers in merge. Table \ref{ab_topK} shows the the performance of different hyper-parameters. (More detail in )

\subsection{Visualization}

Fig \ref{fig:vis} illustrates several HOI detection results. The baseline model tends to focus on semantically similar objects, resulting in lower detection accuracy. In contrast, our model directs attention to the correct interaction regions. Our method also improves localization of human-object pairs and avoids confusion between sibling classes. And Fig \ref{fig:vis_tsne} visualize t-SNE before and after debiasing, showing significantly improved feature distinction across categories

\section{Conclusion}
In this work, we analyze key learning biases in human-object interaction (HOI) detection. To address these biases, we introduce two additional learning objectives. For input-level bias, we employ contrastive and calibration losses. 
To tackle output-level bias, we propose a merge-split learning strategy. Experimental results demonstrate that our approach significantly outperforms the baseline and state-of-the-art methods. 

\noindent\textbf{Acknowledgment:} This work is supported by the National Natural Science Foundation of China (No.62206174).

{
    \small
    \bibliographystyle{ieeenat_fullname}
    \bibliography{main}
}
\maketitlesupplementary

\noindent In Supplementary Material, we provide additional information regarding,

\begin{itemize}
    \item Section~\ref{sec:im_details} Implementation Details
    \begin{itemize}
        \item Section~\ref{sec:an_im} Analysing Implementation Details
        \item Section~\ref{sec:label_prompt} Label Prompt in Details
        \item Section~\ref{sec:HOR_mask} Human-Object Region Mask Attention
    \end{itemize}
    \item Section~\ref{sec:exp_results} More Experimental Results
    \begin{itemize}
        \item Section~\ref{sec:eval_m} Evaluation Metrics
        \item Section~\ref{sec:confuseing_results} Performance on Class Imbalance Categories
        \item Section~\ref{sec:zs} Experimental Result On Zero-shot
        \item Section~\ref{sec:hoiclip} Applying C2C/M2S to HOICLIP
        \item Section~\ref{sec:no_hor_mask} Experimental Result without HOR Mask
        \item Section~\ref{sec:loss_weight} Ablation Studies on Hyper-Parameters
    \end{itemize}
    \item Section~\ref{sec:vis} Qualitative Results
    \begin{itemize}
        \item Section~\ref{sec:compare} Qualitative Comparisons with the Baseline
        \item Section~\ref{sec:cluster_samples} Qualitative Examples for Super Interaction-Object Categories
        % \item Section~\ref{sec:split_samples} Qualitative Examples for the Most Similar  $k$ Categories
    \end{itemize}
\end{itemize}

\section{Implementation Details}
\label{sec:im_details}
In this section, we provide detailed explanation for Figure \textcolor{iccvblue}{2}, implementation details for the transformation \(\mathrm{F_{text}}(\cdot)\) and the human-object region mask attention (HOR mask), as mentioned in Section \textcolor{iccvblue}{3.1.2} and Section \textcolor{iccvblue}{4.3}, respectively.

\subsection{Detailed Explanation for Figure 2}
\label{sec:an_im}

\noindent \textbf{``Sibling'' Identification} 

We define "siblings" at two distinct levels:
\begin{itemize}
    \item \textbf{Input-Level Siblings:} Within a single image, two HOI triplets \texttt{(human, verb, object)} are considered siblings if they share the same object or the same verb (interaction) category.
    
    \item \textbf{Output-Level Siblings:} Two distinct HOI categories are considered siblings if the cosine similarity between their respective classification head weights in the model is high. This similarity serves as a proxy for semantic or feature-space closeness.
\end{itemize}

\noindent \textbf{Sub-figure Analysis}
\begin{itemize}
    \item \textbf{Figure \textcolor{iccvblue}{2}(a) \& (b):} These plots show the training mAP curves for representative sibling HOI pairs. The red lines highlight instances where the learning progress of one category appears to suppress the progress of its sibling, illustrating a competitive effect at both the input (a) and output (b) levels.
    
    \item \textbf{Figure \textcolor{iccvblue}{2}(c):} This bar chart presents the average error rate, computed across all categories, under two conditions. It compares performance on ground-truth instances that have input-level siblings in the same image (red bar) against those that do not (yellow bar). A category's error rate is defined as the proportion of its samples that yield an mAP of 0 during evaluation.
    
    \item \textbf{Figure \textcolor{iccvblue}{2}(d):} This scatter plot visualizes the relationship between a category's final performance and its inherent output-level similarity, focusing on ``non-head classes''. Crucially, the output-level similarity for a category (x-axis) is determined \textit{before training} by calculating the average cosine similarity of its initial classification head weights to those of all other categories. This initial similarity is then plotted against the category's final mAP after training (y-axis).
\end{itemize}

\subsection{Label Prompt in Details}
\label{sec:label_prompt}
In this subsection, we provide additional details about the transformation \(\mathrm{F_{text}}(\cdot)\). As mentioned in Section \textcolor{iccvblue}{3.1.2}, for different inputs, we formalize them into distinct text descriptions, which are then fed into the CLIP~\cite{radford2021learning} text encoder to obtain the corresponding text features. As shown in Table~\ref{tab:text_des}, we introduce the process of generating different text descriptions corresponding to different inputs following GEN-VLKT~\cite{liao2022gen}.

\begin{table}[h]
% \tablestyle{5.0pt}{1.02}
\footnotesize
\begin{tabular}{@{\hskip -0.05ex}c|c}
Input & Text Description  \\
\hline
\hline
object (Object) & \textit{A photo of a/an [Object].}  \\
interaction (Verb) & \textit{A photo of a person [Verb] something.} \\
HOI triplet & \textit{A photo of a person [Verb-ing] a/an [Object].}  \\
\end{tabular}
\caption{{Text descriptions corresponding to different inputs following GEN-VLKT~\cite{liao2022gen}}.}
\label{tab:text_des}
\end{table}

\subsection{Human-Object Region Mask Attention}
\label{sec:HOR_mask}
In this subsection, we provide additional details about the human-object mask attention (HOR mask), as mentioned in Section \textcolor{iccvblue}{4.3}. To guide interaction queries to focus on the correct human-object interaction regions within the image and avoid interference from similar sibling HOIs, we employ an attention mask for the union of each human-object pair. Specifically, for the standard mask attention~\cite{cheng2021mask2former} weights in the cross-attention of $\operatorname{Decoder_{action}}$:
$$
\begin{aligned}
\mathbf{A} & = \operatorname{softmax}\left(\mathcal{M} + W\right), \\
\mathcal{M}_{i}(x, y) & = \begin{cases} 
0 & \text{if } (x, y) \in \operatorname{region}(b_{i}^{{h}}, b_{i}^{{o}}) \\
-\infty & \text{otherwise}
\end{cases},
\end{aligned}
$$
where $\mathbf{A}$ represents the final attention weights, $W$ denotes the initial attention weights computed by $\mathbf{Q}^{a}$ and $\mathbf{V}^{i}$, and $\mathcal{M}_{i}(x, y)$ is the mask applied to the attention weight of the $i$-th interaction query at location $(x, y)$ in the image features. Here, $\operatorname{region}$ is the bounding box that encapsulates both of the given bounding boxes $b_{i}^{{h}}, b_{i}^{{o}}$.

The attention mask effectively filters out most input level ``Toxic Siblings" distractions, enhancing the recognition of HOI triplets. However, it cannot entirely eliminate interference caused by adjacent ``Toxic Siblings" biases. For example, as illustrated in Fig \textcolor{iccvblue}{1 (a)}, a person standing behind a bench and another sitting on the same bench share an overlapping bounding box. To further mitigate input level ``Toxic Siblings" bias, we propose the ``contrastive-then-calibration" (C2C) debiasing objective. The effectiveness and necessity of the C2C debiasing objective are validated in Table \textcolor {iccvblue}{2}.

\section{More Experimental Results}
\label{sec:exp_results}

In this section, we provide additional experimental results, including the performance gains of our method over the baseline on categories affected by class imbalance bias and the model's performance without the HOR mask module. Following LOGICHOI~\cite{li2024neural}, we conducted further ablation experiments on V-COCO~\cite{gupta2015visual} to analyze the effects of loss weights and hyperparameters.

\subsection{Evaluation Metrics}
\label{sec:eval_m}
We use mean Average Precision (mAP) as the primary evaluation metric. A detection is considered a true positive (TP) if (1) the intersection over union (IoU) between the predicted human and object bounding boxes exceeds 0.5, and (2) the predicted action/interaction category is correct. For HICO-DET, mAP is computed in two settings: (i) the default setting, using all test images, and (ii) the Known Object setting, where average precision (AP) is computed for each object on a subset of images containing specific objects. For V-COCO, we use two evaluation settings: Scenario 1, where the detector must report an empty box when the interaction excludes an object, and Scenario 2, where object box detection can be ignored.

\subsection{Performance on bias Categories}
\label{sec:confuseing_results}
\begin{figure}[h]
    % \vspace{-3mm}
    \centering
    \includegraphics[width=0.95\linewidth]{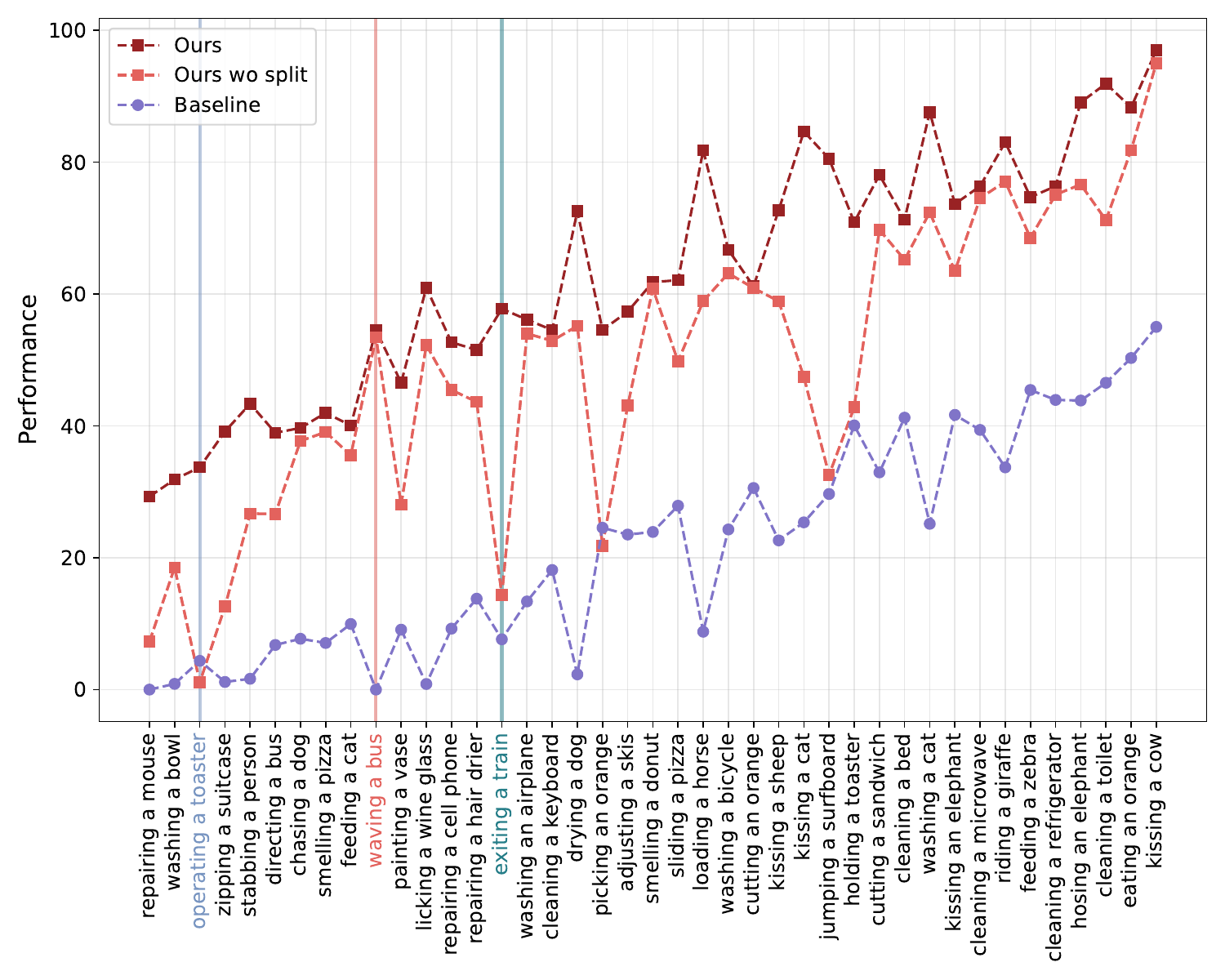}
    \vspace{-2.5mm}
    % \caption{}
            % \vspace{-0.1cm}
    \caption{Experimental results of our method compared to the baseline on categories affected by class imbalance bias.}
    \label{fig:label_gap}
    % \vspace{-5mm}
\end{figure}
In Fig~\ref{fig:label_gap}, we present a comparison of Average Precision (AP) for several HOI categories affected by ``Toxic Siblings" bias. Due to this bias, the baseline model struggles to recognize certain HOI categories, such as \textcolor[HTML]{E3625D}{``waving a bus"} and \textcolor[HTML]{257D88}{``exiting a train"} (see the \textcolor[HTML]{8074C8}{line} in Fig~\ref{fig:label_gap}). 
The merge learning objective partially addresses this by leveraging shared features among similar HOI categories, enabling the recognition of some challenging categories, such as \textcolor[HTML]{E3625D}{``waving a bus"} (AP +55.4\%).
However, ``Toxic Siblings" bias arises not only from long-tail issues but is often exacerbated by shared interactions or objects across categories, making it even more challenging to address. While the merge learning objective facilitates the learning of generalized features, it is insufficient for resolving ambiguities among overly similar categories, such as \textcolor[HTML]{257D88}{``exiting a train"}, and may even intensify confusion between certain categories, such as \textcolor[HTML]{7895C1}{``operating a toaster"}. To overcome this limitation, we introduce the split learning objective, which enables the model to more effectively distinguish between these closely related categories (see the \textcolor[HTML]{992224}{line} in Fig~\ref{fig:label_gap}).

Overall, our combined ``merge-then-split" learning objective effectively mitigates ``Toxic Siblings" bias, resulting in a significant mAP improvement of +18.93\% for HOI categories affected by class imbalance bias.

\subsection{Experimental Result On Zero-shot}
\label{sec:zs}
\begin{table}[h]
    \centering
    \small
    \begin{tabular}{l|c|c|c|c}
    \toprule
    Method & Type & Unseen & Seen & Full \\
    \midrule

    \midrule
      GEN-VLKT & UV & 20.96 &  30.23 & 28.74 \\
      UniHOI & UV & 26.05 & 36.78 & 34.68 \\
      \cellcolor{violet!7}\textbf{Ours} & \cellcolor{violet!7}UV & \cellcolor{violet!7}\textbf{30.47} &\cellcolor{violet!7}\textbf{41.96} &\cellcolor{violet!7}\textbf{38.05} \\

    \midrule
    GEN-VLKT & UO & 10.51 &  28.92 & 25.63 \\
    UniHOI & UO & 19.72 & 34.76 & 31.56 \\
    \cellcolor{violet!7}\textbf{Ours} & \cellcolor{violet!7}UO & \cellcolor{violet!7}\textbf{31.01} &\cellcolor{violet!7}\textbf{41.59} &\cellcolor{violet!7}\textbf{37.23} \\

    \midrule
    GEN-VLKT & NF & 25.05 &  23.38 & 23.71 \\
    UniHOI & NF & 28.45 & 32.63 & 31.79 \\
    SICHOI & NF & 34.52 & 36.06 & 35.75 \\
     \cellcolor{violet!7}\textbf{Ours} & \cellcolor{violet!7}NF & \cellcolor{violet!7}\textbf{36.12} &\cellcolor{violet!7}\textbf{37.58} &\cellcolor{violet!7}\textbf{37.01} \\

    \midrule
    GEN-VLKT & RF & 21.36 &  32.91 & 30.56 \\
    UniHOI & RF & 28.68 & 33.16 & 32.27 \\
    SICHOI & RF & 34.24 & 41.58 & 40.11 \\
     \cellcolor{violet!7}\textbf{Ours} & \cellcolor{violet!7}RF & \cellcolor{violet!7}\textbf{35.48} &\cellcolor{violet!7}\textbf{42.91} &\cellcolor{violet!7}\textbf{40.96} \\

    \midrule

      \bottomrule[1.0pt]
    \end{tabular}
    \caption{Zero-Shot Results On HICO-DET}
    \label{tab:supp_zero}
\end{table}

\noindent Table~\ref{tab:supp_zero} serves as an extension of Table \textcolor{iccvblue}{2}, illustrating the zero-shot performance in greater detail.

\subsection{Applying C2C/M2S to HOICLIP}
\label{sec:hoiclip}
\begin{table}[h]
\centering
\small
\begin{tabular}{lccc}
\toprule
\multirow{2.2}{*}{Method} & \multicolumn{3}{c}{\textbf{HICO-DET (Default)}} \\
\cline{2-4} 
 & Full & Rare & Non-Rare \\
\hline
\midrule
HOICLIP               & 34.54         & 30.71        & 35.70                   \\
                                    
\cellcolor{violet!7}HOICLIP (with C2C and M2S)  & \cellcolor{violet!7}\textbf{43.11} & \cellcolor{violet!7}\textbf{42.69} & \cellcolor{violet!7}\textbf{43.58} \\
\cellcolor{violet!7}                                    &  \cellcolor{violet!7}\textcolor{gray}{+8.42}  &  \cellcolor{violet!7}\textcolor{gray}{+11.57}   &  \cellcolor{violet!7}\textcolor{gray}{+7.84} \\
\bottomrule
\end{tabular}
% }
\caption{Experimental results with applying C2C and M2S to HOICLIP on HICO-DET\!~\cite{chao2018learning} under the default setting.}
\vspace{-3mm}
\label{tab:hoiclip}
\end{table}
\noindent We integrate our C2C and M2S learning objectives into HOICLIP~\cite{ning2023hoiclip}, as shown in Table~\ref{tab:hoiclip}. This increases HOICLIP's HICO-DET mAP to 43.11\% (Full), 42.69\% (Rare) and 43.58\% (Non-Rare), achieving substantial gains of +8. 42\% (Full), +11.57\% (Rare) and +7.84\% (Non-Rare), respectively.

\subsection{Experimental Result without HOR Mask}
\label{sec:no_hor_mask}

\begin{table}[h]
\centering
\small
\begin{tabular}{lccc}
\toprule
\multirow{2.2}{*}{Method} & \multicolumn{3}{c}{\textbf{HICO-DET (Default)}} \\
\cline{2-4} 
 & Full & Rare & Non-Rare \\
\hline
\midrule
Baseline (GEN-VLKT-s)               & 33.75         & 29.25        & 35.10                   \\
ViPLO~\cite{park2023viplo} & 34.95 &33.83 &35.28 \\
ADA-CM~\cite{lei2023efficient} & 38.40 &37.52 &38.66 \\
BCOM~\cite{wang2024bilateral}    &39.34 &39.90 &39.17     \\
\cellcolor{violet!7}Ours (w/o HOR mask)               & \cellcolor{violet!7}{42.01} & \cellcolor{violet!7}{41.55} & \cellcolor{violet!7}{42.21} \\
\cellcolor{violet!7}                                    &  \cellcolor{violet!7}\textcolor{gray}{+2.67}  &  \cellcolor{violet!7}\textcolor{gray}{+1.65}   &  \cellcolor{violet!7}\textcolor{gray}{+3.04} \\
                                    
\cellcolor{violet!7}Ours (with HOR mask)  & \cellcolor{violet!7}\textbf{42.93} & \cellcolor{violet!7}\textbf{42.41} & \cellcolor{violet!7}\textbf{43.11} \\
\cellcolor{violet!7}                                    &  \cellcolor{violet!7}\textcolor{gray}{+3.59}  &  \cellcolor{violet!7}\textcolor{gray}{+2.51}   &  \cellcolor{violet!7}\textcolor{gray}{+3.94} \\
\bottomrule
\end{tabular}
% }
\caption{Experimental results with and without HOR mask on HICO-DET\!~\cite{chao2018learning} under the default setting.}
\vspace{-3mm}
\label{tab:supp_no_hor}
\end{table}

\noindent Table~\ref{tab:supp_no_hor} presents the performance of our method without the HOR mask module, using ResNet-50 (R50) as the backbone and CLIP as the VLM. Notably, the proposed Contrastive-then-Calibration (C2C) and Merge-then-Split (M2S) learning objectives independently mitigate both types of bias, achieving an 8.26\% improvement over the baseline and a performance of 42.01\% (+2.67\% compared to the state-of-the-art BCOM) on the HICO-Det dataset.

\subsection{Ablation Studies on Hyper-Parameters}
\label{sec:loss_weight}

\begin{table}[h]
\centering
 % \small
\begin{tabular}{lcc}
\toprule
 Ablation on & \multicolumn{2}{c}{\textbf{V-COCO}}  \\
% \cline{2-3} 
 loss weights & $AP^{S1}_{role}$ &$AP^{S2}_{role}$  \\
\midrule
\multicolumn{3}{c}{$\lambda_2 = 1, \lambda_3 = 0.5, \lambda_4 = 1, \lambda_5 = 0.5,$ while $\lambda_1=$} \\
\midrule
0.1 & 69.3 & 71.8\\
1 & \textbf{69.8} & \textbf{72.1} \\
10 & 69.5 & 72.0 \\
\midrule
\multicolumn{3}{c}{$\lambda_1 = 1, \lambda_3 = 0.5, \lambda_4 = 1, \lambda_5 = 0.5,$ while $\lambda_2=$} \\
\midrule
0.1 & 69.1 & 71.4 \\
1 & \textbf{69.8} & \textbf{72.1}  \\
10 & 69.2 & 71.6  \\
\midrule
\multicolumn{3}{c}{$\lambda_1 = 1, \lambda_2 = 1, \lambda_4 = 1, \lambda_5 = 0.5,$ while $\lambda_3=$} \\
\midrule
0.05 & 69.5 & 71.7 \\
0.5 & \textbf{69.8} & \textbf{72.1}  \\
5 & 69.4 & 71.8 \\
\midrule
\multicolumn{3}{c}{$\lambda_1 = 1, \lambda_2 = 1, \lambda_3 = 0.5, \lambda_5 = 0.5,$ while $\lambda_4=$} \\
\midrule
0.1 & 69.2 & 71.3 \\
1 & \textbf{69.8} & \textbf{72.1} \\
10 & 69.4 & 71.6 \\
\midrule
\multicolumn{3}{c}{$\lambda_1 = 1, \lambda_2 = 1, \lambda_3 = 0.5, \lambda_4 = 1,$ while $\lambda_5=$} \\
\midrule
0.05 & 69.6 & 71.8 \\
0.5 & \textbf{69.8} & \textbf{72.1} \\
5 & 69.4 & 71.7\\
\bottomrule
\end{tabular}
% \vspace{-3mm}
\caption{Ablation study on loss weights.}
\vspace{-3mm}
\label{tab:loss_weight}
\end{table}

\noindent \textbf{Ablation on loss weights.} To evaluate the effectiveness of the proposed learning objectives, we conducted ablation experiments by varying their respective weights, as summarized in Table ~\ref{tab:loss_weight}. Specifically, $\lambda_1$ denotes the weight of the original detection loss, while $\lambda_2$, $\lambda_3$, $\lambda_4$, and $\lambda_5$ represent the weights of the contrastive loss $\mathcal{L}_\text{con}$, calibration loss $\mathcal{L}_\text{cal}$, merge loss $\mathcal{L}_\text{merge}$, and split loss $\mathcal{L}_\text{split}$, respectively.

\begin{table}[h]
\centering
 % \small
\begin{tabular}{lcc}
\toprule
 Ablation on & \multicolumn{2}{c}{\textbf{V-COCO}}  \\
% \cline{2-3} 
 \# of $k_1$ and $k_2$ & $AP^{S1}_{role}$ &$AP^{S2}_{role}$  \\
\midrule
$k_1=1,k_2=10$ & 69.0 & 71.2 \\
$k_1=2,k_2=10$ & \textbf{69.8} & \textbf{72.1} \\
$k_1=3,k_2=10$ & 69.5 & 71.9 \\
$k_1=5, k_2=10$ & 69.3 & 71.6 \\

\bottomrule
\end{tabular}
% \vspace{-3mm}
\caption{Ablation study on $k_1$.}
\vspace{-3mm}
\label{tab:supp_hp}
\end{table}

\noindent \textbf{Ablation on $k_1$.} As a supplement to Table \textcolor{iccvblue}{3} in the main text, we conduct further ablation experiments on the hyper-parameter $k_1$ mentioned in Section \textcolor{iccvblue}{3.3.2}. The corresponding results are shown in Table~\ref{tab:supp_hp}.

\section{Qualitative Results}
\label{sec:vis}

\subsection{Qualitative Comparisons with the Baseline}
\label{sec:compare}
\begin{figure*}[t]
\centering
\vspace{-3mm}
\includegraphics[width=0.97\textwidth]{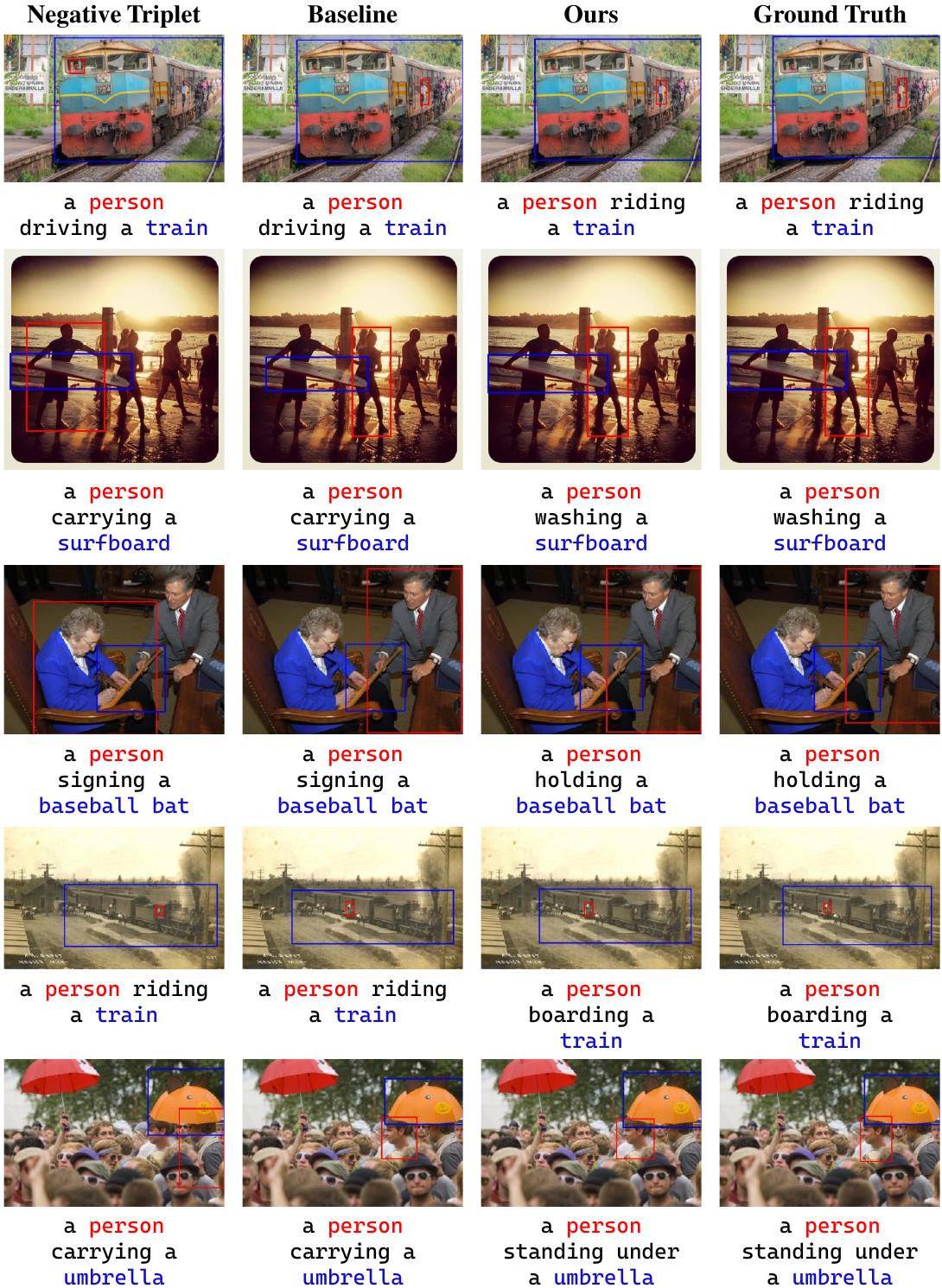}
% \vspace{-3mm}
\caption{{Qualitative comparisons of the first type of input level ``Toxic Siblings" bias between our method and the baseline.}}
\label{fig:compare_bias1_1}
\vspace{-3mm}
\end{figure*}

\begin{figure*}[t]
\centering
% \vspace{-3mm}
\includegraphics[width=1\textwidth]{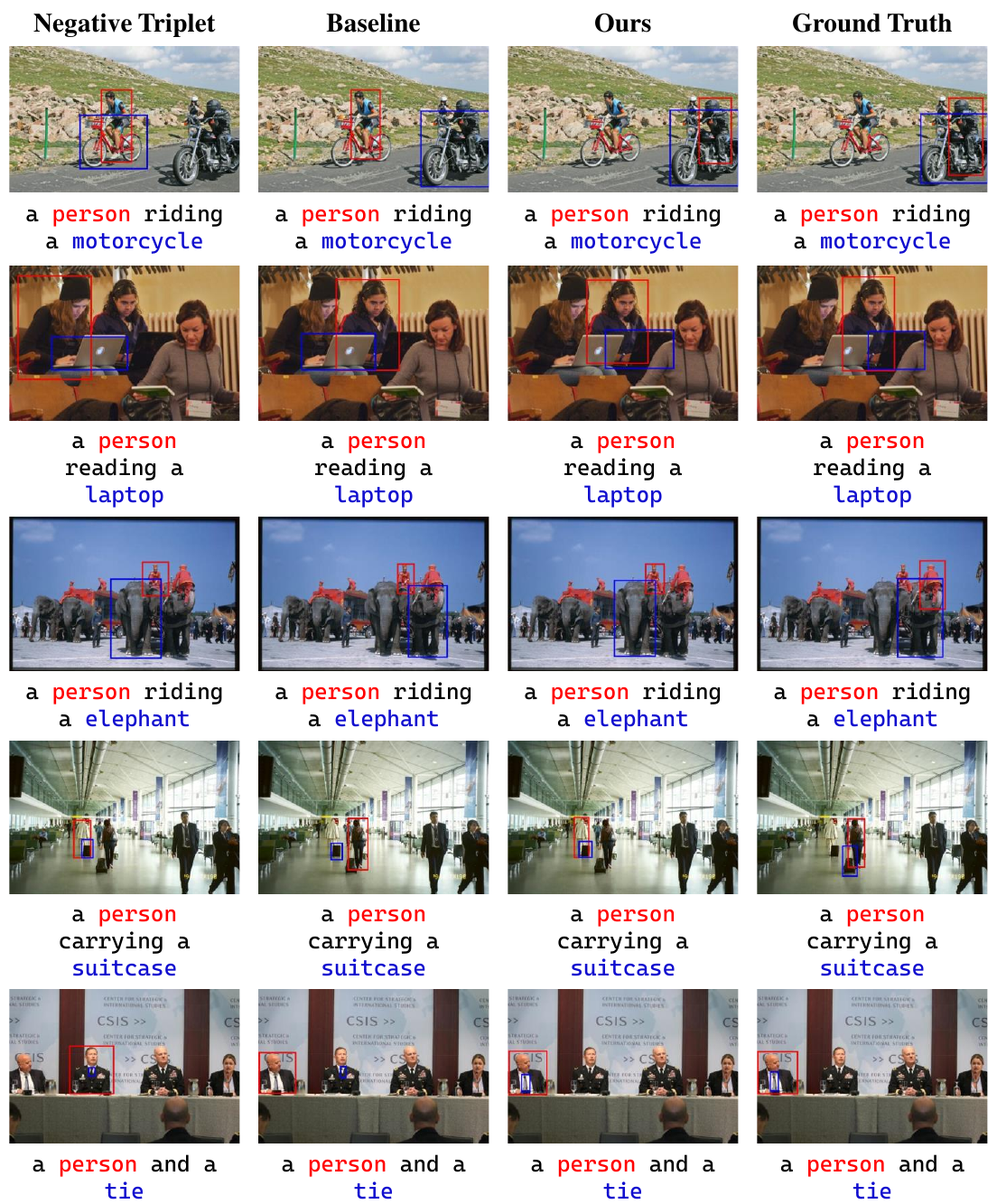}
% \vspace{-7mm}
\caption{{Qualitative comparisons of the second type of input level ``Toxic Siblings" bias between our method and the baseline.}}
\label{fig:compare_bias1_2}
\vspace{-3mm}
\end{figure*}

\noindent \textbf{Qualitative Comparisons of input level Bias.} Fig~\ref{fig:compare_bias1_1} and ~\ref{fig:compare_bias1_2} present qualitative comparison results between the baseline~\cite{liao2022gen} and our model under the influence of input level ``Toxic Siblings" bias on HICO-DET~\cite{chao2018learning}. From left to right, the images depict the negative triplet causing bias, the predictions of the baseline, our predictions, and the ground truth.

Fig~\ref{fig:compare_bias1_1} illustrates the first type of input level ``Toxic Siblings" bias: 
interaction actions are misclassified due to the influence of nearby, similar negative triplets (see column 1), leading to incorrect predictions (see column 2). 
For instance, in the first row, the ground truth (see column 4) is ``a person riding a train". However, due to the presence of a similar negative HOI ``a person driving a train” (see column 1), which shares the same object ``train" with the ground truth, the baseline misclassifies the interaction as ``driving", even though the human and object are correctly predicted (see column 2). 
By identifying negative triplets in the surrounding context and differentiating them through the proposed C2C learning objective, our model accurately predicts the interaction (see column 3).

Fig~\ref{fig:compare_bias1_2} illustrates the second type of input level ``Toxic Siblings" bias: 
unrelated and unpaired humans and objects are mistakenly predicted to have interactions due to the influence of nearby triplets (see column 1 and 4) that share the same human or object in the context. 
For example, in the first row, the baseline incorrectly matches a black motorcycle with a person wearing a blue vest (see column 2), influenced by the nearby interactions of ``a person riding a motorcycle". 
The proposed C2C learning objective guides the model’s attention to the correct spatial regions, mitigating the effects of input level ``Toxic Siblings" bias and enabling accurate predictions (see column 3).

\begin{figure*}[t]
\centering
\vspace{-3mm}
\includegraphics[width=0.95\textwidth]{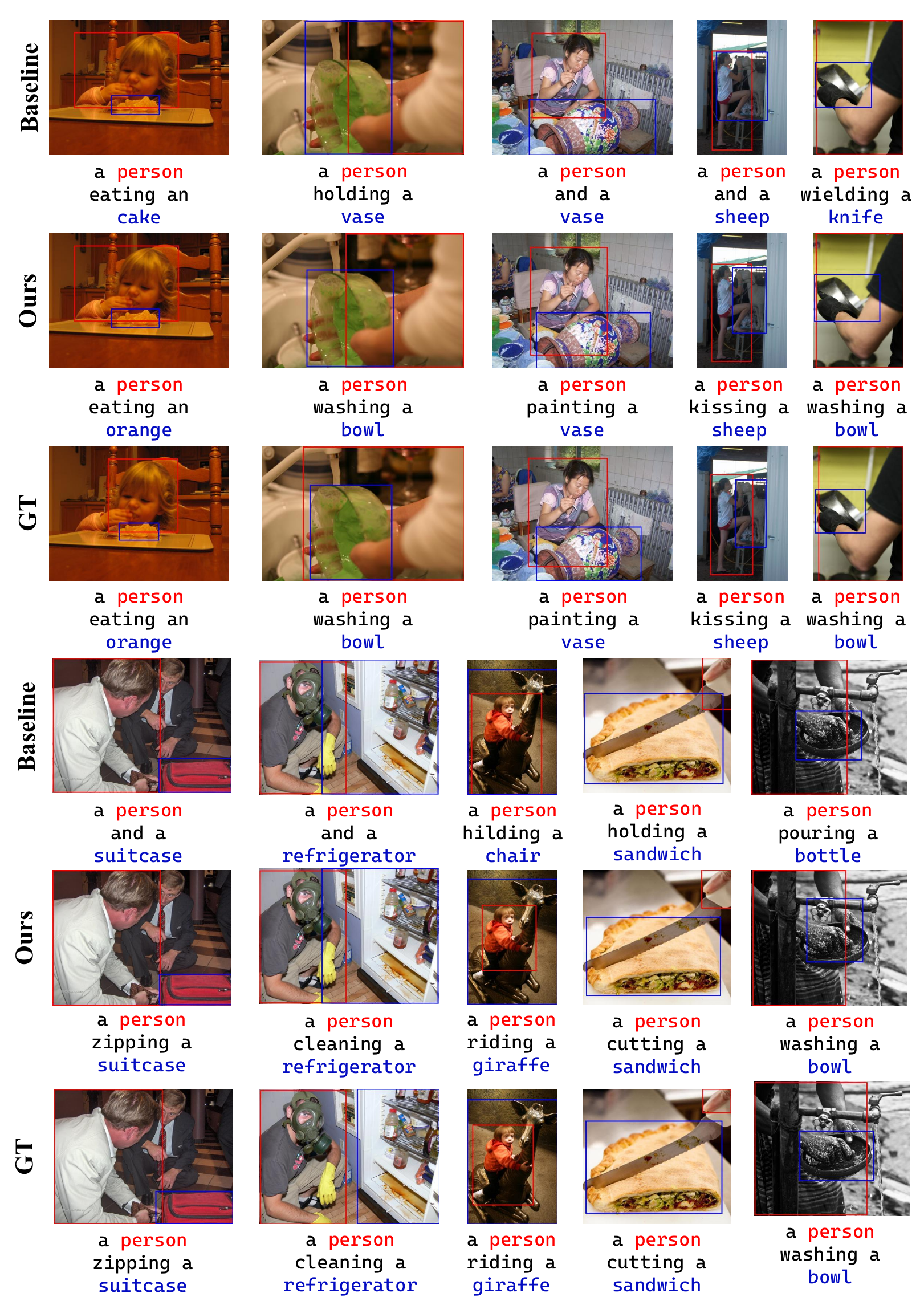}
% \vspace{-3mm}
\caption{{Qualitative comparisons of output level ``Toxic Siblings" bias between our method and the baseline.}}
\label{fig:compare_bias2_1}
\vspace{-3mm}
\end{figure*}

\noindent \textbf{Qualitative Comparisons of Class Imbalance Bias.} 
The qualitative results in Fig~\ref{fig:compare_bias2_1} demonstrate the effectiveness of our Merge-then-Split (M2S) learning objective in mitigating class imbalance bias. From top to bottom, the rows represent the baseline predictions, our model’s predictions, and the ground truth.
Due to output level ``Toxic Siblings" bias, the baseline tends to misclassify long-tail HOI categories, such as ``eating an orange" or ``zipping a suitcase" into semantically similar sibling head classes, like ``eating a cake" or ``and a suitcase" (see rows 1 and 4). In contrast, our debiased model successfully predicts the correct long-tail HOI categories (see rows 2 and 5). This improvement is attributed to the M2S learning objective, which effectively reduces misclassifications for long-tail classes.

\subsection{Qualitative Examples for Super Interaction-Object Categories}
\label{sec:cluster_samples}
Tables~\ref{tab:clu1} and ~\ref{tab:clu2} present selected category clusters obtained through the merge process. The clusters in Table~\ref{tab:clu1} primarily encompass actions involving the manipulation of everyday objects using hands or mouth, while those in Table~\ref{tab:clu2} focus on interactions that reflect the diverse forms of emotional contact, interaction, and specific actions between humans and animals or other individuals. Table~\ref{tab:clu3} illustrates the multimodal interaction behaviors between humans and food objects through actions such as eating, inspecting, and smelling.

\begin{table}[t]
% \vspace{2em}
% \footnotesize
\centering
\resizebox{0.3\textwidth}{!}{
\begin{tabular}{c}
    \toprule
    HOI categories in Super Category 1 \\
\midrule
    \textit{holding a bowl, stirring a bowl} \\
 \textit{ washing a bowl, licking a bowl} \\
 
 \textit{ cutting a cake, holding a cake} \\
 
 \textit{ drinking with a cup, holding a cup} \\
 
 \textit{ pouring a cup, sipping a cup} \\
 
 \textit{ washing a cup, holding a pizza}  \\
 
 \textit{ holding a remote, cutting a sandwich} \\
 
    \bottomrule
\end{tabular}
}
\vspace{-0.2em}
\caption{Super-category of manipulating everyday objects using hands or mouth.}
\label{tab:clu1}
\end{table}

\begin{table}[t]
% \vspace{2em}
% \footnotesize
\centering
\resizebox{0.3\textwidth}{!}{
\begin{tabular}{c}
    \toprule
    HOI categories in Super Category 2 \\
\midrule
    \textit{ holding a cow, hugging a cow} \\

 \textit{ kissing a cow, holding a dog} \\
 
 \textit{ hosing a dog, hugging a dog} \\
 
 \textit{  kissing a dog, washing a dog} \\
 
 \textit{ holding a horse, hugging a horse} \\
 
 \textit{ kissing a horse, washing a horse}  \\
 
 \textit{ holding a person, hugging a person} \\
 
 \textit{ kissing a person, stabbing a person} \\
 
    \bottomrule
\end{tabular}
}
\vspace{-0.2cm}
\caption{Super-category of interactions between humans and animals.}
\vspace{-0.5cm}

\label{tab:clu2}
\end{table}

\begin{table}[t]
% \vspace{2em}
% \footnotesize
\centering
\resizebox{0.3\textwidth}{!}{
\begin{tabular}{c}
    \toprule
    HOI categories in Super Category 3 \\
\midrule
    \textit{ eating an apple, inspecting an apple} \\
 \textit{ smelling an apple,  and an apple} \\
 \textit{ eating a banana, inspecting a banana} \\
 \textit{  smelling a banana, and a banana} \\
 \textit{ eating a broccoli, smelling a broccoli} \\
 \textit{ and a broccoli, eating a carrot}  \\
 \textit{ smelling a carrot, and a carrot} \\
 \textit{ eating a donut, smelling a donut} \\
 \textit{ and a donut, eating a hot dog} \\
 \textit{ and a hot dog, eating an orange} \\
 \textit{ inspecting an orange, and an orange} \\
 
    \bottomrule
\end{tabular}
}
% \vspace{0.5em}
\caption{Super-category of interaction between humans and food objects.}
\label{tab:clu3}
\end{table}

\clearpage

\end{document}